\documentclass[acmlarge,screen,nonacm]{acmart}

\AtBeginDocument{%
  }

\usepackage{amsmath}
\usepackage{array}
\usepackage{booktabs}
\usepackage{textcomp}
\usepackage{graphicx}
\usepackage{enumitem}
\usepackage{lineno}

\makeatletter
\renewcommand{\@secfont}{\sffamily\bfseries\large\section@raggedright}
\renewcommand{\@subsecfont}{\sffamily\bfseries\large\section@raggedright}
\renewcommand{\@subsubsecfont}{\sffamily\bfseries}
\newcommand{\subsubsubsection}{\def\@toclevel{4}%
  \@startsection{paragraph}{4}{\z@}%
  {-.5\baselineskip \@plus -2\p@ \@minus -.2\p@}%
  {-3.5\p@}%
  {\ACM@NRadjust{\@subsubsecfont\@adddotafter}}}
\newcommand{\subsubsubsubsection}{\def\@toclevel{5}%
  \@startsection{subparagraph}{5}{\z@}%
  {-.25\baselineskip \@plus -1\p@ \@minus -.1\p@}%
  {-3.5\p@}%
  {\ACM@NRadjust{\sffamily\bfseries\@adddotafter}}}
\makeatother

\makeatletter
\if@ACM@nonacm
  \renewcommand{\@ACM@checkaffil}{%
    \if@ACM@instpresent\else
      \ClassWarningNoLine{acmart}{No institution present for an affiliation}%
    \fi
  }
\fi
\ExplSyntaxOn
\seq_new:N \l_usti_affiliations_seq
\prop_new:N \l_usti_affiliation_numbers_prop
\cs_new_protected:Npn \USTI@resetaffiliations {
  \seq_clear:N \l_usti_affiliations_seq
  \prop_clear:N \l_usti_affiliation_numbers_prop
}
\cs_new_protected:Npn \USTI@addaffiliation #1 {
  \prop_get:NnNTF \l_usti_affiliation_numbers_prop {#1} \l_tmpa_tl
    { }
    {
      \seq_put_right:Nn \l_usti_affiliations_seq {#1}
      \tl_set:Nx \l_tmpa_tl {\seq_count:N \l_usti_affiliations_seq}
      \prop_put:NnV \l_usti_affiliation_numbers_prop {#1} \l_tmpa_tl
    }
  \tl_gput_right:Nx \@currentauthors {
    \exp_not:N \textsuperscript {\exp_not:V \l_tmpa_tl}
  }
}
\cs_new_protected:Npn \USTI@printaffiliations {
  \seq_map_indexed_inline:Nn \l_usti_affiliations_seq {
    \int_compare:nNnT {##1} > {1} {,\space}
    \@ACM@resetaffil
    \textsuperscript{##1}##2\@ACM@checkaffil
  }
}
\ExplSyntaxOff
\renewcommand{\@mkauthors@i}{%
  \def\@currentauthors{}%
  \USTI@resetaffiliations
  \def\and{}%
  \def\@author##1{%
    \ifx\@currentauthors\@empty
      \gdef\@currentauthors{\@authorfont\MakeUppercase{##1}}%
    \else
      \g@addto@macro\@currentauthors{,\space\MakeUppercase{##1}}%
    \fi
  }%
  \def\email##1##2{}%
  \def\affiliation##1##2{\USTI@addaffiliation{##2}}%
  \def\institution##1{\global\@ACM@instpresenttrue##1\ignorespaces}%
  \global\setbox\mktitle@bx=\vbox{%
    \noindent\unvbox\mktitle@bx\par\medskip
    \addresses
    \noindent\@currentauthors\par
    {\@affiliationfont\noindent\USTI@printaffiliations\par}%
    \medskip
  }%
}
\makeatother

\begin{document}


\title{Towards Spatio-Temporal Intelligence Systems in Urban Space: A Twin-Agent Perspective}

\author{Zhen Dong}
\affiliation{\institution{Wuhan University}}
\email{dongzhenwhu@whu.edu.cn}

\author{Haiping Wang}
\affiliation{\institution{The Hong Kong University of Science and Technology}}
\correspondingauthor
\email{hpwang@ust.hk}

\author{Wei Tu}
\affiliation{\institution{Shenzhen University}}
\email{tuwei@szu.edu.cn}

\author{Yuan Liu}
\affiliation{\institution{The Hong Kong University of Science and Technology}}
\email{yuanly@ust.hk}

\author{Bing Wang}
\affiliation{\institution{The Hong Kong Polytechnic University}}
\email{bingwang@polyu.edu.hk}

\author{Xuan Sun}
\affiliation{\institution{Nankai University}}
\email{sunxuan@nankai.edu.cn}

\author{Jinzhou Cao}
\affiliation{\institution{Shenzhen Technology University}}
\email{caojinzhou@sztu.edu.cn}

\author{Ronald Clark}
\affiliation{\institution{University of Oxford}}
\email{ronald.clark@cs.ox.ac.uk}

\author{Zhiheng Fu}
\affiliation{\institution{The Hong Kong Polytechnic University}}
\email{zhiheng.fu@polyu.edu.hk}

\author{Mohammed Bennamoun}
\affiliation{\institution{The University of Western Australia}}
\email{mohammed.bennamoun@uwa.edu.au}

\author{Danfeng Hong}
\affiliation{\institution{Southeast University}}
\email{danfeng.hong@seu.edu.cn}

\author{Yuanzheng Cui}
\affiliation{\institution{Chinese Academy of Sciences}}
\email{yzcui@niglas.ac.cn}

\author{Gemine Vivone}
\affiliation{\institution{National Research Council of Italy}}
\email{gemine.vivone@cnr.it}

\author{Pedram Ghamisi}
\affiliation{\institution{University of Iceland}}
\email{p.ghamisi@hzdr.de}

\author{Bisheng Yang}
\affiliation{\institution{Wuhan University}}
\email{bshyang@whu.edu.cn}

\author{Qingquan Li}
\affiliation{\institution{Shenzhen University}}
\email{liqq@szu.edu.cn}

\author{Deren Li}
\affiliation{\institution{Wuhan University}}
\email{drli@whu.edu.cn}

\authorsaddresses{
Emails: Zhen Dong (\href{mailto:dongzhenwhu@whu.edu.cn}{dongzhenwhu@whu.edu.cn}), 
Haiping Wang (Corresponding author, \href{mailto:hpwang@ust.hk}{hpwang@ust.hk}),
Wei Tu (\href{mailto:tuwei@szu.edu.cn}{tuwei@szu.edu.cn}),
Yuan Liu (\href{mailto:yuanly@ust.hk}{yuanly@ust.hk}),
Bing Wang (\href{mailto:bingwang@polyu.edu.hk}{bingwang@polyu.edu.hk}),
Xuan Sun (\href{mailto:sunxuan@nankai.edu.cn}{sunxuan@nankai.edu.cn}),
Jinzhou Cao (\href{mailto:caojinzhou@sztu.edu.cn}{caojinzhou@sztu.edu.cn}),
Ronald Clark (\href{mailto:ronald.clark@cs.ox.ac.uk}{ronald.clark@cs.ox.ac.uk}),
Zhiheng Fu (\href{mailto:zhiheng.fu@polyu.edu.hk}{zhiheng.fu@polyu.edu.hk}),
Mohammed Bennamoun (\href{mailto:mohammed.bennamoun@uwa.edu.au}{mohammed.bennamoun@uwa.edu.au}),
Danfeng Hong (\href{mailto:danfeng.hong@seu.edu.cn}{danfeng.hong@seu.edu.cn}),
Yuanzheng Cui (\href{mailto:yzcui@niglas.ac.cn}{yzcui@niglas.ac.cn}),
Gemine Vivone (\href{mailto:gemine.vivone@cnr.it}{gemine.vivone@cnr.it}),
Pedram Ghamisi (\href{mailto:p.ghamisi@hzdr.de}{p.ghamisi@hzdr.de}),
Bisheng Yang (\href{mailto:bshyang@whu.edu.cn}{bshyang@whu.edu.cn}),
Qingquan Li (\href{mailto:liqq@szu.edu.cn}{liqq@szu.edu.cn}),
Deren Li (\href{mailto:drli@whu.edu.cn}{drli@whu.edu.cn})}

\renewcommand{\shortauthors}{Dong et al.}

\begin{teaserfigure}
  \centering
  \includegraphics[width=\textwidth]{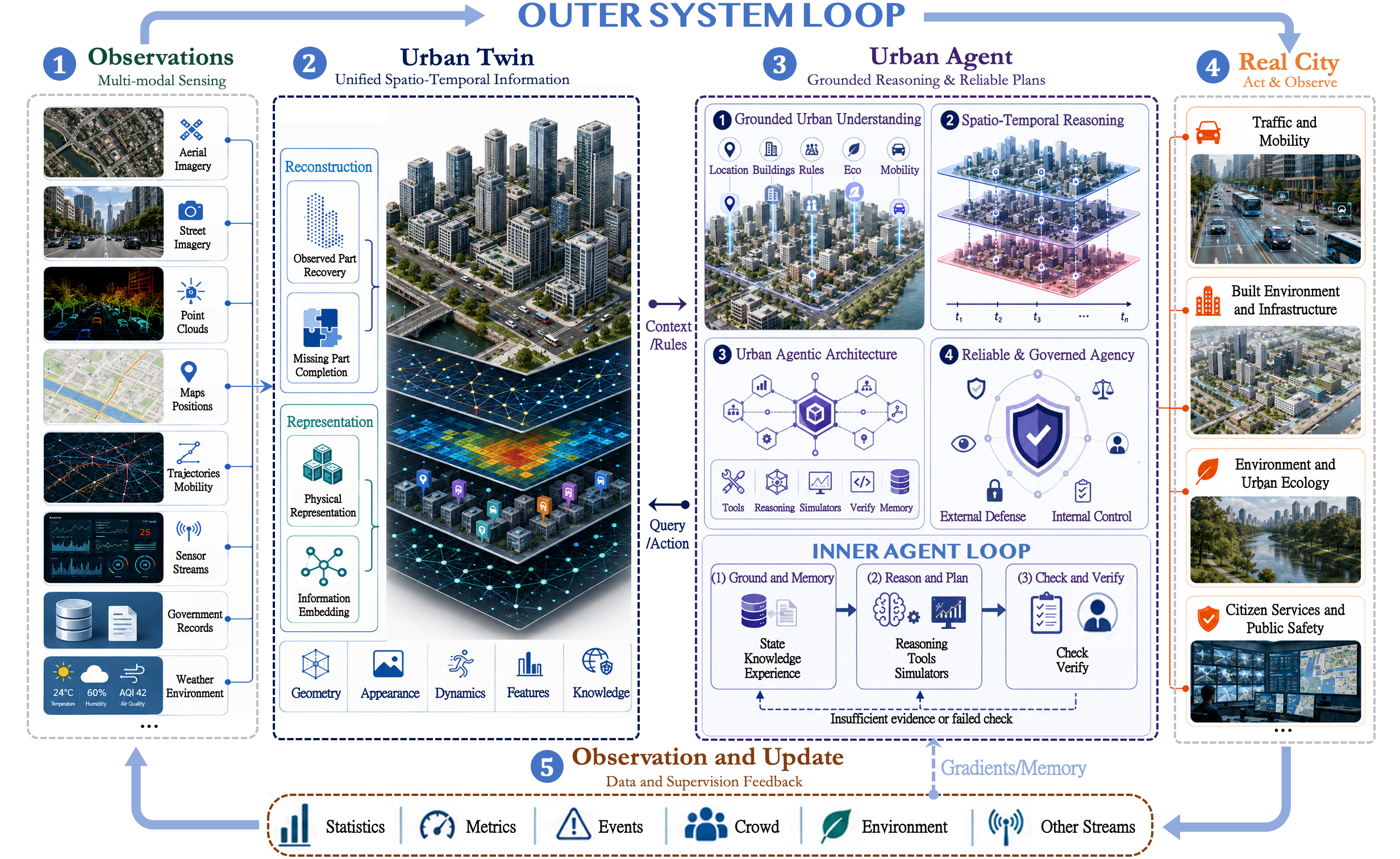}
  \caption{Scope of this review. With the twin-agent perspective, we review the technologies for constructing an urban twin, developing grounded urban agent, and integrating both components into a future STI system for urban domains.}
  \label{fig:review_scope}
\end{teaserfigure}

\begin{abstract}
\textbf{ABSTRACT.} 
Cities must coordinate domains like infrastructure, mobility, environmental management, emergency response, and public services as conditions evolve. An urban Spatio-Temporal Intelligence (STI) system helps the city monitor citywide states, analyze problems, plan and execute interventions, and adapt through feedback, making urban governance more timely, effective, and transparent. Recent urban twinning and agentic AI techniques provide foundations for such systems. From a twin-agent perspective, this survey takes a system-level view of how these disparate techniques can be coherently integrated into an urban STI system. Specifically, we review how the twin reconstructs and represents a persistent, information-enriched urban environment, and how the agent understands the twin via interfaces, reasons over it, acts reliably, and continuously evolves with the city. Built upon this synthesis, we summarize a two-loop blueprint, where an inner agent loop for grounding, reasoning, and verification, and an outer system loop linking data, twin, agentic action, feedback, and update. We then use this to structure an STI research agenda and show how a shared prototype supports diverse urban domains.
\end{abstract}

\maketitle

\section{I. Introduction}

Urban operations follow a recurring cycle: cities observe changing conditions, decide how to respond, and assess the effects of their actions~\cite{batty2013new,liu2015social}. This cycle underpins urban domains such as infrastructure maintenance, mobility management, environmental planning, emergency response, and public services. Yet manual workflows and isolated domain systems usually draw on narrow data and perspectives, yielding delayed and ineffective responses to increasingly complex citywide demands~\cite{zhang2025metacity}.

Building an urban Spatio-Temporal Intelligence (STI) system is the key to meeting such demands~\cite{li2026perspectives}. Such a system should unify these fragmented data and perspectives, consistently track evolving city states, reason across states and knowledge, optimize plans, apply timely and effective actions, and improve continuously through urban feedback.

\noindent \textbf{Twin-agent perspective.} In our previous Neural City work~\cite{dong2026neural}, we framed this system through a two-stage twin-agent architecture. The \textit{twin} maintains a computable and evolving account of the city, while the \textit{agent} grounds understanding, reasoning, and action in that state. Their coupling turns fragmented urban evidence into an operational context for city-scale analysis and intervention.

\begin{table*}[h]
  \centering
  \caption{Scope and technical depth across related reviews. ``Concept'' indicates an explicit framing, whereas ``Techniques'' indicates comprehensive coverage of its key subtopics, technical approaches, and implementation details. ``$\checkmark$'' marks comprehensive coverage at the indicated level; ``--'' marks the absence of such coverage.}
  \vspace{-10pt}
  \label{tab:review_comparison}
  \small
  \setlength{\tabcolsep}{5pt}
  \resizebox{\textwidth}{!}{%
  \begin{tabular}{lcccccc}
    \toprule
    Reviews & \multicolumn{2}{c}{Twin} & \multicolumn{2}{c}{Agent} & \multicolumn{2}{c}{System loop} \\
    \cmidrule(lr){2-3}\cmidrule(lr){4-5}\cmidrule(lr){6-7}
      & Concept & Techniques & Concept & Techniques & Concept & Techniques \\
    \midrule
    Twin Reviews~\cite{lehtola2022digital,xu2024leveraging,jeddoub2025data,sajadieh2025simulation,cao2025reconstructing,kong20253d}
      & $\checkmark$ & $\checkmark$ & -- & -- & -- & -- \\
    Agent Reviews~\cite{dorobantu2026geospatial,han2025large,hashemi2026comprehensive,liang2025foundation,zhang2024urban,zhang2025towards,zhang2025metacity,shao2025spatial}
      & -- & -- & $\checkmark$ & $\checkmark$ & -- & -- \\
    System Reviews~\cite{xu2026towards,ye2026towards}
      & $\checkmark$ & -- & $\checkmark$ & -- & $\checkmark$ & -- \\
    \textbf{This review}
      & $\checkmark$ & $\checkmark$ & $\checkmark$ & $\checkmark$ & $\checkmark$ & $\checkmark$ \\
    \bottomrule
  \end{tabular}%
  }
\end{table*}

\noindent \textbf{Related reviews.} 
As shown in Table \ref{tab:review_comparison}, existing reviews clarify individual parts of urban STI, but generally stop at their component boundaries. 
Reviews of twins cover reconstruction\cite{xu2024leveraging,cao2025reconstructing,kong20253d} and  lifecycles~\cite{lehtola2022digital,jeddoub2025data,sajadieh2025simulation}, while leaving agentic decision-making outside their scope.  
On the agent side, reviews cover the core foundation models~\cite{liang2025foundation,zhang2024urban,zhang2025towards,zhang2025metacity,shao2025spatial}, spatial reasoning~\cite{dorobantu2026geospatial}, and agentic systems~\cite{han2025large,hashemi2026comprehensive}. Across these reviews, urban states are generally treated as raw observations rather than persistent twins.
    
Recent papers connect twins and agents through multi-agent orchestration~\cite{xu2026towards} or human-AI co-learning~\cite{ye2026towards}. However, they emphasize system visions or applications and provide limited detail on how the two subsystems are constructed and coupled through state interfaces and feedback.

\begin{figure*}
  \centering
  \includegraphics[width=\textwidth]{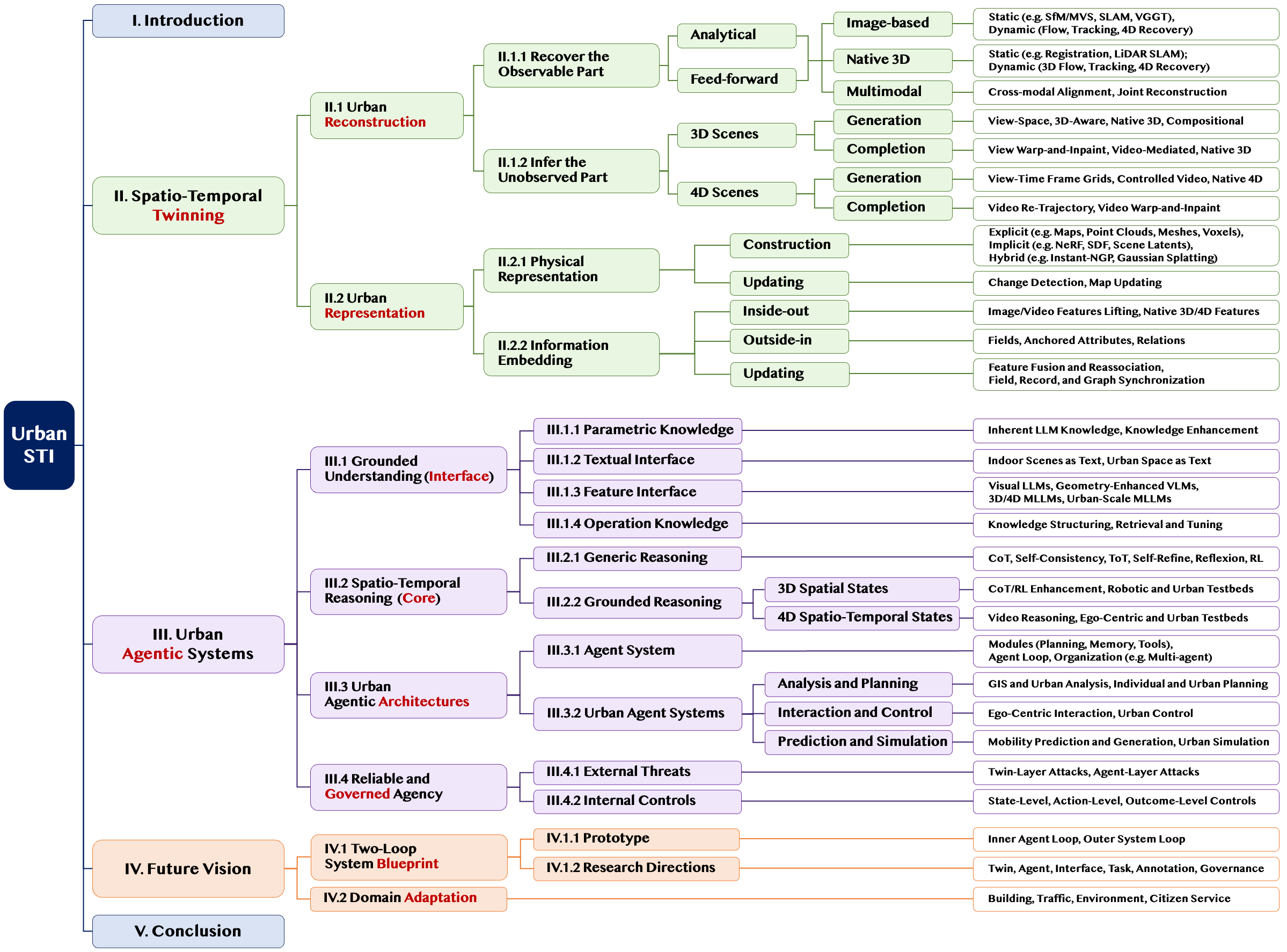}
  \caption{Detailed technical organization of this review. Urban twinning reconstructs, completes, represents, and enriches city states; urban agents access these states through textual and feature interfaces, reason over them, and operate through governed agentic architectures; the future vision couples both subsystems through a two-loop blueprint and adapts it across urban domains. Leaf labels summarize closely related technical topics.}
  \label{fig:review_detail}
\end{figure*}

\noindent \textbf{Scope and contributions.} Our central contribution is to recast fragmented twin and agent views as a joint urban STI system, as summarized in Fig.~\ref{fig:review_scope}. 
We organize the two components as internally \textit{coherent} and operationally \textit{coupled} subsystems. The twin constructs and maintains a persistent, operable urban state rich in information, while the agent interprets, reasons over, and acts reliably on that state. Their interface makes twin states readable by agents, while feedback returns observed outcomes to the twin and the wider system. 

This synthesis yields a two-loop blueprint adaptable across urban domains. Beyond this system coherence, as shown in Fig.~\ref{fig:review_detail}, we provide a detailed \textit{technical} account of how each subsystem is built, from twin reconstruction and representation to agent grounding, reasoning, architecture, and governance, together with their applications and limitations.

\noindent \textbf{Context organization.} Section II traces how urban observations are reconstructed and represented as operable urban twins. Section III examines how agents access these states, reason over them, and operate reliably in urban environments. Section IV brings both components together into an STI system, outlines its research agenda, and considers its adaptation across urban domains. Section V concludes the review.

\section{II. Spatio-Temporal Twinning}

\begin{figure*}[t]
  \centering
  \includegraphics[width=\textwidth]{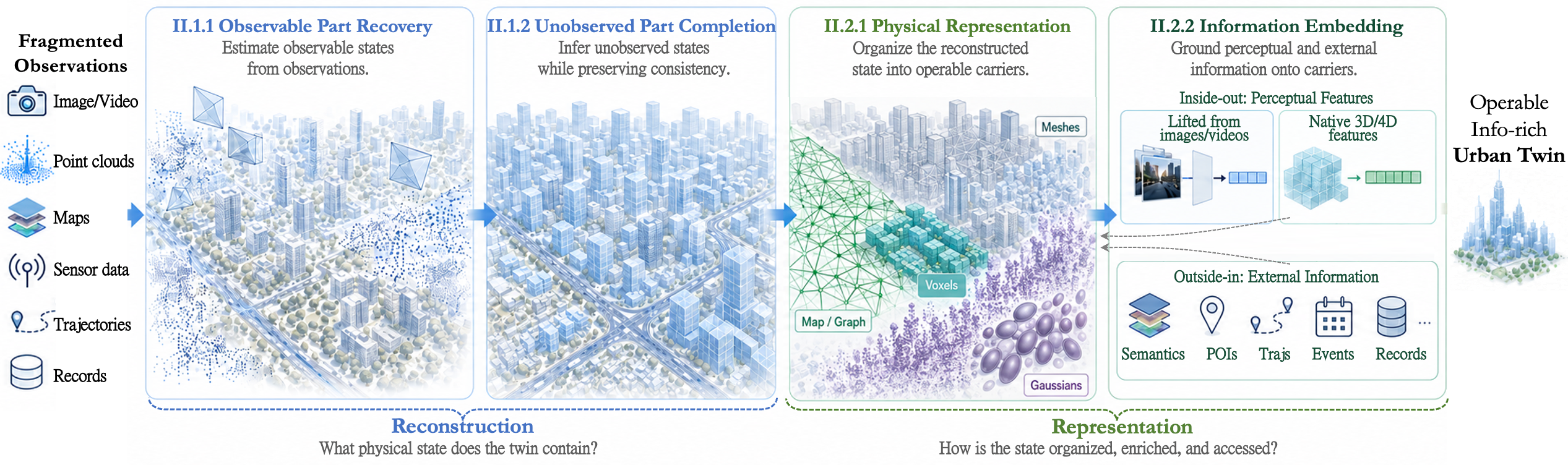}
  \caption{The four-stage urban twinning framework. Recovery and completion determine the physical state contained in the twin, while physical representation and information embedding organize and enrich that state for subsequent operation.}
  \label{fig:twinning_vision}
\end{figure*}

\noindent\textbf{A Vision for Twinning.} Existing reviews typically organize urban scene twinning around individual technical problems, such as 3D/4D reconstruction\cite{cao2025reconstructing}, completion and generation\cite{roldao20223d,wen20253d}, or 3D representation learning\cite{guo2019deep,du2026foundation}. 
These reviews cover the individual components well, but give less attention to how they form an operational twin. We organize their connections through the four-stage framework in Fig.~\ref{fig:twinning_vision}.

First, \textit{recovery} estimates scene geometry, appearance, camera states, and motion directly constrained by observations. Second, \textit{completion} infers unobserved spatial and temporal states while preserving consistency with the observed scene. Third, \textit{physical representation} organizes this reconstructed state into persistent and operable carriers. Finally, \textit{information embedding} grounds perceptual features and external urban knowledge onto these carriers. The first two stages constitute \textit{reconstruction}, determining what physical state the twin contains; the latter two constitute \textit{representation}, determining how that state is organized, enriched, and accessed.

\subsection{II.1 Urban Reconstruction}

Urban reconstruction builds the physical counterpart through two complementary tasks. Metric reconstruction recovers geometry and motion supported by observations, while physical completion infers unobserved spatial and temporal states.

\subsubsection{II.1.1 Recover the Observable Part}
\begin{figure*}[t]
  \centering
  \includegraphics[width=\textwidth]{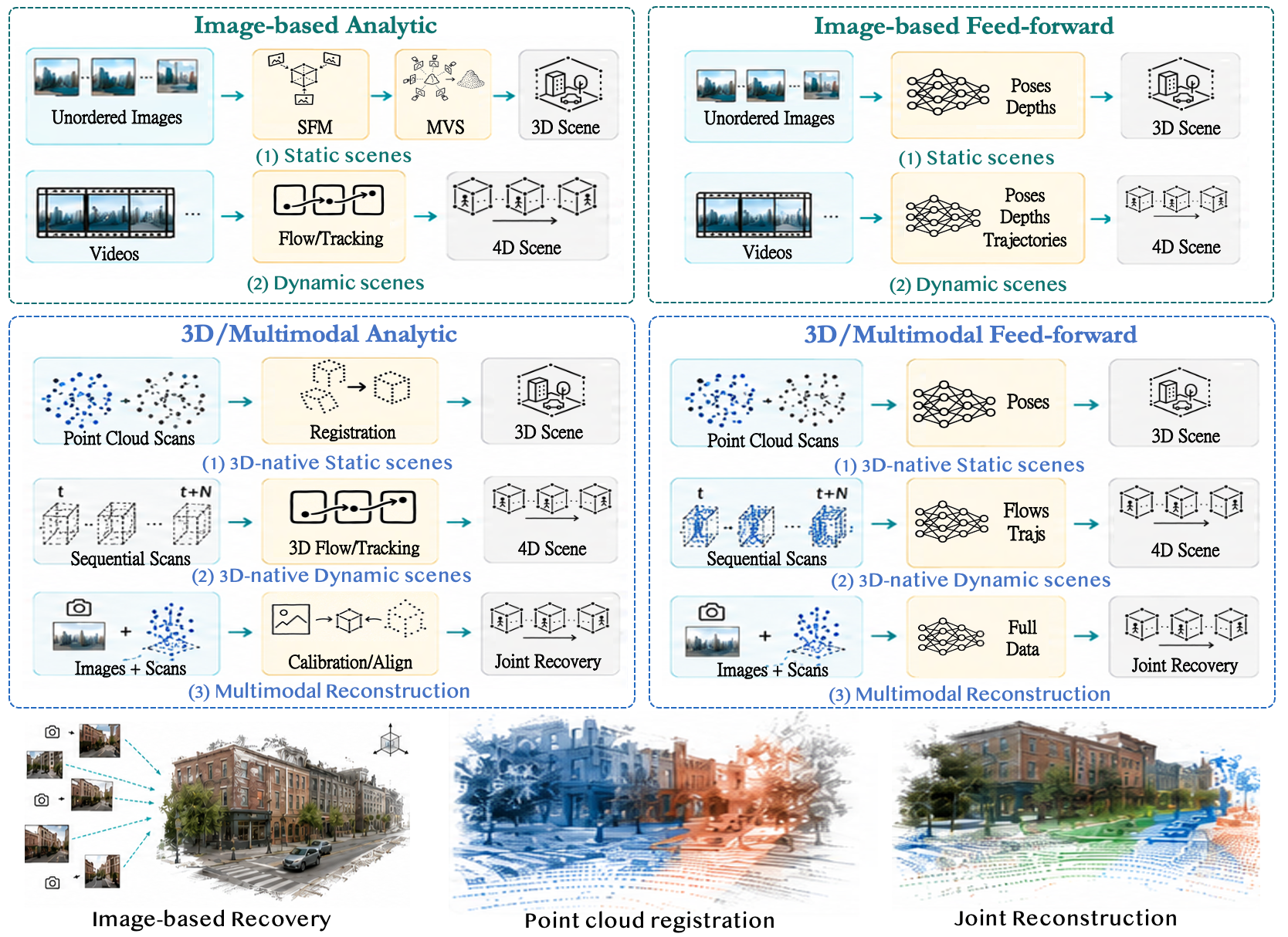}
  \caption{Recovering observable urban states. Analytical and feed-forward paradigms process images, 3D data, or multimodal observations to recover static or dynamic scenes.}
  \label{fig:recover_observed}
\end{figure*}

Given large scale observations such as images, videos, and point clouds, this part focuses on estimating urban geometry and motion directly constrained by measured evidence. Existing systems mainly follow two paradigms, specifically analytical reconstruction and feed-forward reconstruction, as organized in Fig.~\ref{fig:recover_observed}.

\subsubsubsection{II.1.1.1 Analytical Methods} 

They cast reconstruction as a inverse problem. Given observations $\mathbf{y}$, they recover a scene state by solving $\hat{\mathbf{z}}=\arg\min_{\mathbf{z}}E(\mathbf{z};\mathbf{y})$, where $\mathbf{z}$ may include camera poses, geometry, apperance, and motion, and $E$ encodes the observation and prior constraints. The state is optimized separately for each scene, typically through iterative or staged estimation.

\subsubsubsubsection{Image-based Reconstruction}
Early work mostly uses images to reconstruct the static and dynamic scenes.

\textbf{(1) Static Scenes.}
A bunch of works accept an unordered image collection\cite{irschara2007towards,agarwal2011building} to solve the static 3D scene. They first extract handcrafted local features\cite{lowe2004distinctive} or learned features\cite{detone2018superpoint,liu2019gift,revaud2019r2d2}, and then establish correspondences across views through sparse feature matching\cite{sarlin2020superglue,leroy2024grounding} or dense matching\cite{sun2021loftr}. Structure-from-Motion (SfM)\cite{schonberger2016structure} estimates camera poses and sparse 3D points through epipolar geometry and triangulation on correspondences, where Bundle Adjustment (BA)\cite{triggs1999bundle,agarwal2010bundle} jointly refines the cameras and 3D points by minimizing reprojection error. Given the recovered cameras, Multi-View Stereo (MVS) estimates framewise dense depth maps or surfaces through cross-view triangulation under texture consistency\cite{seitz2006comparison,wang2021multi} or feature cost volume regression\cite{yao2018mvsnet,yao2019recurrent,zhang2023dsc}. Recent neural fields\cite{mildenhall2021nerf,wang2021neus} or Gaussian Splatting methods\cite{kerbl20233d,huang20242d} fit scene geometry together with appearance by optimization, as further introduced in Sec.II.2.1.1.

Another line processes long, structured videos captured along moving trajectories\cite{cornelis20083d}. One line assumes a static scene and focuses on camera localization and spatial mapping. Visual Odometry (VO)\cite{nister2004visual,fraundorfer2012visual} sequentially estimates relative motion from matches between consecutive frames and integrates the estimates into a camera trajectory. Visual SLAM\cite{mur2015orb,teed2021droid,zhu2022nice} jointly estimates camera poses and a persistent spatial map by matching, keyframe selection and local BA. Loop closure and pose graph optimization further correct accumulated drift\cite{taketomi2017visual}. When additional sensors are available, IMU measurements provide gravity and motion priors\cite{qin2018vins}, while GNSS observations anchor the trajectory to a global reference frame\cite{liu2021optimization}.

\textbf{(2) Dynamic Scenes.}
Videos of dynamic scenes mix camera changes with scene motion. Analytical methods recover scene dynamic states at three levels.

\textit{(a) Flow.} Optical flow estimates a dense 2D displacement field between adjacent frames from brightness constancy and spatial smoothness\cite{beauchemin1995computation,horn1981determining}, subsequently strengthened by robust penalties\cite{black1993framework}, gradient constancy\cite{brox2004high}, and descriptor matching for large displacements\cite{brox2010large,weinzaepfel2013deepflow}. With calibrated stereo\cite{huguet2007variational,ma2019deep} or multiview videos\cite{basha2013multi,pons2007multi}, temporal correspondence can be coupled with cross-view geometry to recover 3D scene flow, whose image projection induces optical flow\cite{vedula1999three}. These formulations recover local motion fields but do not by themselves maintain persistent identities over long sequences.

\textit{(b) Tracking.} Tracking aims to recover long-term trajectories within videos. It first builds cross-frame correspondences by matching selected feature points\cite{tomasi1993shape}, object patches\cite{comaniciu2003kernel}, or learned features\cite{tumanyan2024dino}, and then consolidates noise or missing matches into temporally consistent trajectories.
Particle Video\cite{sand2008particle} explicitly optimizes semi-dense long-range trajectories under appearance and neighborhood deformation constraints. DINO-Tracker\cite{tumanyan2024dino} instead jointly adapts pretrained DINO features and a tracker to each test video for refined matching and tracking. Neural fields encode correspondence in a shared continuous spacetime representation. Neural Trajectory Fields\cite{wang2021neural} refines a smooth DCT-parameterized trajectory for each spacetime point, while OmniMotion\cite{wang2023tracking} optimizes a quasi 3D canonical volume from enhanced pairwise correspondences, producing globally consistent dense tracks through occlusion. These formulations recover persistent trajectories without necessarily reconstructing a complete 4D scene.

\textit{(c) Dynamic reconstruction.} Beyond motion recovery, dynamic reconstruction integrates geometry, appearance, and scene motion into a persistent 4D representation. Flow and tracks often serve as correspondence constraints or supervision. 
Early dynamic neural fields extend radiance field queries from space to spacetime, mapping position, time, and viewing direction to time-varying density and color\cite{pumarola2021d,wang2021neural}, while dynamic Gaussian methods optimize Gaussian primitives and their temporal deformation or trajectories\cite{wu20244d,wang2025shape}. Besides joint optimization, Street Gaussians\cite{yan2024street} decomposes the environment into a static background and tracked dynamic objects; recent FunRec\cite{delitzas2026funrec} tracks and segments movable parts or objects by clustering to reconstruct functional 3D scenes for interaction.
Broader treatments of analytical methods for flow, tracking, and 4D reconstruction are available in \cite{zhai2021optical,smeulders2013visual,cao2025reconstructing}.

Although the image-based methods above can recover scene representations with rich appearance, their geometry may be incomplete or distorted in weakly textured areas or under illumination changes\cite{martin2021nerf}, and metric scale may be unavailable\cite{schonberger2016structure}.

\subsubsubsubsection{3D Native Reconstruction}
Direct 3D observations from LiDAR or depth cameras provide accurate metric surface samples\cite{yang2024ubiquitous}, shifting reconstruction from inferring geometry from images to registering and integrating geometric frames.

\textbf{(1) Static Scenes.}
Static reconstruction from 3D observations is primarily a registration problem that aligns partial scans in a common frame\cite{huang2021comprehensive}. Typical pipelines establish point correspondences with geometric descriptors\cite{choy2019fully,ao2021spinnet,ao2023buffer}, robustly estimate pairwise rigid transformations\cite{wang2023roreg,zhang20233d,qin2022geometric}, refine them with Iterative Closest Point (ICP) algorithm\cite{besl1992method}, and enforce multiview consistency through pose graph synchronization\cite{arrigoni2016spectral,gojcic2020learning,wang2023robust}. For long streams, LiDAR SLAM performs continuous state estimation through scan-to-map registration\cite{xu2021fast,xu2022fast,pan2021mulls}, with loop closure and backend optimization reducing accumulated drift and maintaining global map consistency\cite{pan2024pin,zhang20243d}. Broader reviews are available in \cite{huang2023cross,huang2021comprehensive}.

\textbf{(2) Dynamic Scenes.}
These methods accept sequential 3D observations recording dynamic scenes. 

\textit{(a) Flow} estimates metric displacement between adjacent scans. For instance, RGBD-SF\cite{quiroga2014dense} jointly optimize color and depth correspondences under semi-rigid motion constraints. ICP-Flow\cite{lin2024icp} instead associates locally rigid objects across LiDAR scans and recovers their transformations with ICP. 

Then, \textit{(b) Tracking} maintains long-range 3D correspondences. For instance, EulerFlow\cite{vedder2025neural} estimates a continuous spacetime ODE, integrating local pairwise motion to long-range point trajectories. Lepard\cite{li2022lepard} improves feature matching between deformable scans for motion tracking and dynamic ICP modeling. Some mothods\cite{yi2025moving,jang2023toss} jointly conduct moving object segmentation and temporal association in urban LiDAR streams. 

\textit{(c) Dynamic reconstruction} further integrates dynamic measurements into persistent geometry. Some early work jointly optimizes deformation tracking and surface fusion of non-rigid objects\cite{zollhofer2014real,newcombe2015dynamicfusion}.
For scenes, under a rigid object assumption, RigidFusion\cite{long2021rigidfusion} jointly segments, tracks, and reconstructs the static background and a large dynamic rigid object, while CoFusion\cite{runz2017co} maintains separate 3D models for multiple independently moving objects. At the field level, 4D implicit mapping\cite{zhong20243d} fits a time-dependent Truncated Signed Distance Function to LiDAR sequences to reconstruct a consistent map without segmentation. Broader discussions are available in \cite{deng2022survey,guan2025multi}.
Direct 3D observations provide metric scale and geometry, but reconstruction remains sensitive to low overlap, geometric degeneracy, and is typically textureless\cite{huang2023cross,zhang20243d,deng2022survey}.

\subsubsubsubsection{Multimodal Reconstruction}
Combined image and 3D observations integrate rich appearance with accurate metric geometry, but must first be spatially aligned. For independently acquired data, the modality gap complicates cross-modal feature construction\cite{wang2021p2,wang2024freereg,hong2026foundation} and reliable dense correspondence estimation\cite{ren2022corri2p,kang2024cofii2p}, which in turn affects pose estimation such as PnP\cite{lepetit2009ep}. In synchronized sensor rigs, extrinsic calibration estimates a fixed cross-sensor transformation\cite{li2023automatic,liao2023se,iyer2018calibnet}. Once registered, photometric, geometric, and pose constraints can be jointly optimized to recover textured geometry\cite{pan2025pings,whelan2015elasticfusion}.

\noindent \textbf{Overall}, analytical methods offer interpretable objectives, but remain sensitive to challenging observations like low overlap or textureless areas, and the optimization is computationally expensive at scale. COLMAP, for example, may take hours to reconstruct thousands of images\cite{schonberger2016structure,yang2025fast3r}.

\subsubsubsection{II.1.1.2 Feed-forward Methods} 
Feed-forward methods accelerate reconstruction with learned priors, recovering scene states through forward inference rather than optimizing each scene from scratch\cite{cao2025reconstructing}.

\subsubsubsubsection{Image-Based Reconstruction}
This branch progresses from recovering static geometry from individual images or static videos to estimating motion and consistent 4D structure from videos.

\textbf{(1) Static Scenes.} Feed-forward image reconstruction first matured in monocular depth estimation\cite{bhat2023zoedepth}. With vision transformers\cite{dosovitskiy2020image} and increasingly large image datasets with 3D annotations\cite{ling2024dl3dv,wang2026spatialvid}, models achieve transferable relative depth\cite{yang2024depth,ke2024repurposing} and metric depth\cite{yin2023metric3d,bochkovskiy2025depth,wang2026moge}. The target then expands from one image to stereo reconstruction\cite{wen2025foundationstereo,wang2024dust3r,jang2025pow3r}, and further to joint camera and geometry (pixel-aligned depth or point cloud) recovery from multiple views\cite{wang2026pi,wang2025vggt,lin2025depth,zhang2025flare}. Dense cross-image attention recovers globally consistent point or depth maps, camera poses, and even instance or semantic features\cite{li2025iggt,fan2024large,li2025urbansam,sun2026uni3r} in one pass, trading some metric precision for robustness and speed\cite{wang2026pi,wang2026vggt}.

At large scales, joint multiview transformers incur quadratic cross-view attention, motivating sparse scene graphs\cite{elflein2025light3r}, sparse or linear attention\cite{zhang2026lite3r}, KV compression\cite{elflein2026vgg}, and parallel processing\cite{yang2025fast3r} to reduce memory and improve throughput. 
Specifically, for long videos with ordered frames, feed-forward predictions are often implemented through chunking or history compression. 
Chunking methods split the long video into overlapping short clips for separately lightweight prediction and align them in a common frame\cite{liu2025slam3r,deng2025vggt,maggio2026vggt}, but local alignment errors can accumulate as drift. 
Online incremental methods instead compress historical frames to compact memories for subsequent lightweight attention interaction. CUT3R\cite{wang2025continuous} compresses history into a recurrent state, although states with a fixed length may forget or collapse over long sequences\cite{chen2026ttt3r}. Later systems mitigate this bottleneck through confidence-weighted state updates\cite{chen2026ttt3r}, multi-timescale retention\cite{cheng2026horizonstream}, or separate camera and geometry memories\cite{liu2026mem3r}.
Others externalize scene content as cached spatial features\cite{wang20253d} or trajectory contexts\cite{chen2026geometric}, as well as featured keypoints\cite{wu2026point3r,chen2025long3r} or voxels\cite{wang2026amb3r}, allowing memory to slightly grow with the reconstruction to reduce information loss.

\textbf{(2) Dynamic Scenes.}
Here, methods also progress from short flow and long tracking to 4D reconstruction.

\textit{(a) Flow} networks predict dense displacement between adjacent frames, progressing from convolutional pyramids with local warping and cost volumes\cite{dosovitskiy2015flownet,sun2018pwc} to recurrent all pairs matching and cost volume embedding\cite{teed2020raft,huang2022flowformer}.

\textit{(b) Tracking} extends the temporal horizon and preserves identity across time. Localized trackers propagate selected objects or small masks\cite{chen2022visual,bertinetto2016fully,bhat2019learning,cui2022mixformer,ravi2025sam}. Query-based models maintain arbitrary points across longer clips\cite{harley2022particle,doersch2023tapir,karaev2024cotracker,guerrier2026pointst3r}. Dense variants approach trajectories for every pixel\cite{neoral2024mft}. With estimated depth and camera motion, 2D tracks can be lifted into world space trajectories\cite{xiao2024spatialtracker,xiao2025spatialtrackerv2}. 

\textit{(c) Dynamic reconstruction} recovers temporally coherent geometry from videos of dynamic scenes, with methods differing in whether object motion is explicitly predicted.
Implicit dynamics models do not explicitly output scene motion, such as trajectories or velocities, but predict only geometry and cameras at each timestep. They use dynamic masks\cite{li2025megasam} or motion-aware attention\cite{fang2026more,hu2026vggt4d} to suppress motion interference in reconstruction, or auxiliary flow cues to supervise or coherently align geometric predictions\cite{zhang2025monst3r,lu2025align3r,fei2024driv3r}. When needed, motion must be recovered subsequently by establishing correspondences across the frame predictions.
Explicit dynamics models instead predict geometry together with motion, representing how points move across time as dynamic point maps\cite{sucar2025dynamic,sucar2026v}, 2D/3D trajectories or spacetime correspondences\cite{feng2025st4rtrack,zhang2026efficiently,wang20254d,lu2025trackingworld}, or velocity vectors\cite{he2026dynamicvggt}.

\subsubsubsubsection{3D Native Reconstruction}
For {\textit{static scenes}}, recent methods register unposed point cloud scans in a single forward pass. FUSER\cite{jiang2026fuser} directly predicts scan global poses by the model and refines them through diffusion in $\mathrm{SE}(3)$ space. Register Any Point\cite{pan2025register} instead transports the input points into a registered scene by point cloud diffusion, from which the view poses are recovered.
For {\textit{dynamic scenes}}, 3D Flow networks predict metric displacement between successive RGB-D frames\cite{teed2021raft} or point clouds\cite{liu2019flownet3d,puy2020flot,wu2020pointpwc,mittal2020just,gojcic2021weakly}. Beyond flow, long-range 3D tracking still focuses mainly on segmenting moving LiDAR objects\cite{chen2021moving,mersch2022receding} or detecting and associating rigid objects into trajectories\cite{yin2021center}. Early attempts at long-range tracking of arbitrary points across 3D data sequences use recurrent architectures\cite{fan2019pointrnn}, neural field fitting\cite{wang2022neural}, or multimodal tracking\cite{rajivc2025multi}, but the challenge remains open\cite{wang2024sequential}.

\subsubsubsubsection{Multimodal Reconstruction}
Feed-forward fusion of image and point cloud observations first emerged in RGB-guided depth completion, where an aligned image completes a partial depth map\cite{tan2026masked,wang2025depth}. MapAnything\cite{keetha2026mapanything} and OmniVGGT\cite{peng2026omnivggt} extend this paradigm to multiview reconstruction, jointly processing heterogeneous but incomplete collections of images, depth maps, and camera priors to recover geometry in a common scene.

\noindent \textbf{Overall}, feed-forward methods provide efficient, scalable, and robust reconstruction through learned end-to-end inference. Their limited interpretability and weaker long-range geometric consistency, however, leave analytical optimization important for fine-grained metric accuracy\cite{maggio2026vggt}. Hybrid systems that combine both paradigms may therefore provide a stronger balance of speed and accuracy\cite{chen2026geometric}.

\subsubsection{II.1.2 Infer the Unobserved Part}

\begin{figure*}[t]
  \centering
  \includegraphics[width=\textwidth]{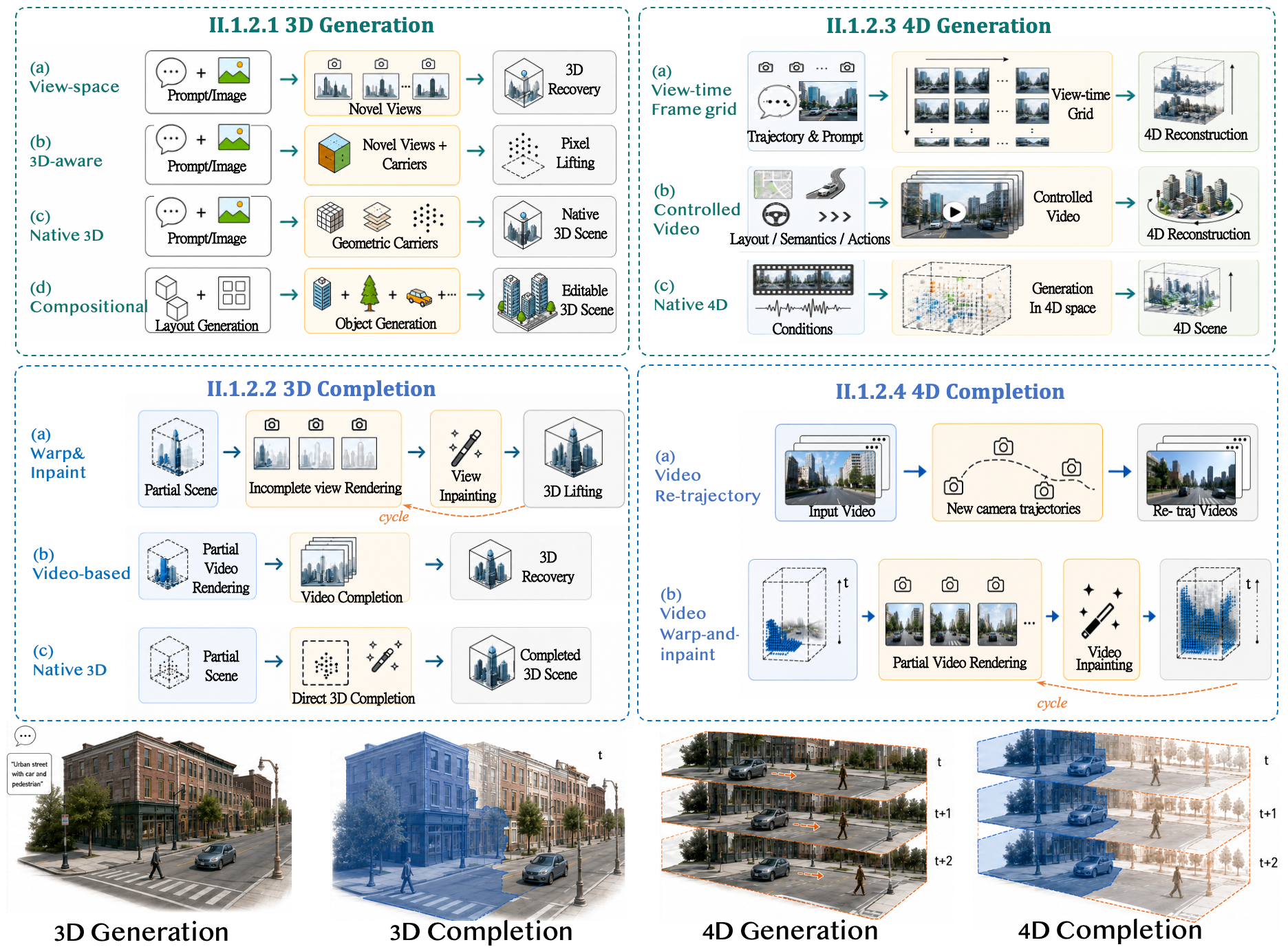}
  \caption{Inferring unobserved scene states. Generation synthesizes new 3D scenes or 4D evolution from weak conditions, whereas completion extends partial scene observations while preserving consistency with the measured spatial and temporal evidence.}
  \label{fig:infer_unobserved}
\end{figure*}

Physical completion infers unobserved spatial structures, appearance, and temporal states from incomplete states. Unlike reconstruction, which mainly estimates what is directly constrained by sensor evidence, completion asks how far a system can extrapolate beyond that evidence while still remaining faithful to the particular observed scene\cite{roldao20223d,xu20233d}.
Recent completion methods increasingly draw on advances in generative modeling\cite{wen20253d}, making the boundary between completion and generation less distinct. We therefore review both, mainly at scene scope, as shown in Fig.~\ref{fig:infer_unobserved}.

\subsubsubsection{II.1.2.1 3D Generation} 
Generation synthesizes new 3D content from prompts, images, or other conditions. Early 3D scene generation relied heavily on procedural rules and handcrafted pipelines, offering controllability but limited diversity and realism\cite{parish2001procedural,muller2006procedural,fisher2012example,wen20253d}. 

Inspired by developments in image diffusion\cite{chen2025comprehensive}, video diffusion\cite{melnik2024video}, and interactive world models\cite{bruce2024genie,lu2025genex,team2026advancing}, recent methods increasingly use generative priors to produce 3D content.

\textit{(a) View-space generation.} Methods such as CAT3D\cite{gao2024cat3d}, Stable Virtual Camera\cite{zhou2025stable}, and NVComposer\cite{li2025nvcomposer} synthesize novel views under a given camera trajectory from single or sparse inputs. The generated views can then provide additional evidence for reconstructing the underlying 3D scene.

\textit{(b) 3D-aware generation.} This route couples view generation with explicit 3D signals or carriers. WVD\cite{zhang2025world} jointly generates RGB and XYZ frames, while Wonderland\cite{liang2025wonderland}, Prometheus\cite{yang2025prometheus}, and Bolt3D\cite{szymanowicz2025bolt3d} predict multiview and pixel-aligned 3DGS carriers for fast scene level generation.

\textit{(c) Native 3D generation.} These methods generate directly within 3D representations. DiffIndScene\cite{ju2024diffindscene} synthesizes indoor TSDF-based meshes through diffusion over 3D voxels. Sat2Scene\cite{li2024sat2scene}, Sat2City\cite{hua2025sat2city} and Sat3DGen\cite{qian2026sat3dgen} map satellite observations to 3D city distributions through 3D voxel- or triplane-based diffusion. GaussianCity\cite{xie2025generative} and ScenDi\cite{guo2026scendi} directly generate 3D Gaussian primitives for novel scenes.
Tile-based methods\cite{engstler2025syncity,chen2025trellisworld} construct coherent large scale scenes by treating each city tile as a self-contained local world and jointly optimizing neighboring tiles for seamless integration.

\textit{(d) Compositional generation.} A complementary direction combines asset and layout generation to support embodied interaction and digital twin composition\cite{li2026realtwin}. For instance, SAM 3D\cite{chen2026sam} produces editable assets from segmented image regions; PhysX-Anything\cite{cao2026physx} further targets physical assets, and graph or layout priors then arrange the above generated assets into coherent and editable scenes\cite{lin2024instructscene}.

\subsubsubsection{II.1.2.2 3D Completion} In contrast, completion extends a measured but incomplete 3D representation while preserving consistency with its observed regions. 
Early spatial completion filled small missing samples through local interpolation\cite{ku2018defense} or surface fitting\cite{sorkine2004least,davis2002filling}. Later learning methods expanded the target from local surfaces to geometric scene volumes\cite{dai2018scancomplete}, densified point distributions\cite{park2023implicit,kwon2022implicit,zhang2025win,yang2024tulip}, and semantic occupancy fields\cite{song2017semantic,roldao2020lmscnet,rist2021semantic,wei2023surroundocc,tian2023occ3d,wang2023openoccupancy,zhang2023occformer}. However, these methods still capture scene geometry only coarsely and provide limited texture modeling.
Recent generative completion focuses on complating realistic scene distributions, allowing novel view synthesis and following three main families. 

\textit{(a) View-wise warp-and-inpaint.} One family iteratively renders incomplete new viewpoints from a partial scene, applies RGB/depth inpainting to the partial rendering\cite{lei2023rgbd2}, and lifts the completed area back into the 3D scene for extension\cite{chung2023luciddreamer,wang2025vistadream,chen2025scenecompleter}. 

\textit{(b) Video-mediated completion.} Another family renders a partial scene as an incomplete visual or feature video, uses video diffusion to complete it, and then reconstructs or fuses the generated observations back into 3D\cite{yu2024viewcrafter,ren2025gen3c,liu2026reconx,cao2025mvgenmaster,chen2024mvsplat360}. 

\textit{(c) Native 3D completion.} A smaller line directly uses 3D diffusion optimization to complete spatial structure in 3D space\cite{zheng2026self}. 
Broader analyses of 3D scene generation and completion are provided in recent surveys\cite{wen20253d,tang2025recent}.

\subsubsubsection{II.1.2.3 4D Generation} Compared with 3D generation, which focuses on cross-view geometric and appearance consistency, 4D generation must additionally coordinate scene dynamics. Video diffusion provides the dominant generative prior\cite{melnik2024video,brooks2024video,wan2025wan}.

\textit{(a) View-time frame grid generation.} A representative formulation organizes 4D content as a structured view-time frame grid\cite{wu2025cat4d}, whose generated multi-view and multi-time frames provide dense evidence for downstream 4D reconstruction\cite{wu20244d}. Given single or sparse observations, these methods fill the grid by coordinating fixed-time multiview slices with fixed-view temporal slices\cite{zhao2025genxd,wang20254real,sun2024dimensionx}. Scaling this grid further motivates curated 3D/4D data construction\cite{zhao2025genxd,wu2025cat4d} and sparse attention for efficient cross-grid communication\cite{kuang2024collaborative}.

\textit{(b) Spacetime controlled video generation.} Beyond appearance coherence, this line constrains video generation by structured layouts, physical attributes, and actions.
In driving scenarios, 3D bounding boxes\cite{chen2025unimlvg}, HD maps\cite{alhaija2025cosmos,ren2025cosmos}, and semantic voxels\cite{lu2025infinicube} control the layout of generated traffic videos. More generally, 3D tracking videos provide a unified interface for camera control, motion transfer, and object manipulation\cite{gu2025diffusion}. 
Dynamic G-buffer\cite{liang2025diffusionrenderer} sequences condition video generation on editable physical properties like  geometry, material, and lighting channels. 
Actions directly specify scene evolution.
Recent world models\cite{team2026advancing,zhu2025aether} use motion actions to control video camera trajectory evolution while requiring the underlying 4D scene to remain consistent. 
Physical methods instead treat actions as interactions with the scene, using simulator rollouts to condition the trajectories or deformations of objects in generated videos and thereby preserve the physical plausibility of scene evolution\cite{li2025wonderplay,wang2026physctrl}.
These generated videos can subsequently be converted into explicit 4D representations through downstream reconstruction\cite{wu20244d}.

\textit{(c) Native 4D generation.} Rather than treating video as the endpoint, recent methods directly generate explicit 4D states. For example, DynamicCity\cite{bian2025dynamiccity} produces large-scale 4D semantic occupancy, Aether\cite{zhu2025aether} and 4DNeX\cite{chen20254dnex} extend video generation to predict pixel-aligned depths or dynamic point clouds, and Diff4Splat\cite{pan2026diff4splat} maps video diffusion to dynamic Gaussian Splatting generation. Further details are available in recent reviews\cite{kong20253d,miao2025advances}.

\subsubsubsection{II.1.2.4 4D Completion} Earlier temporal completion extrapolated beyond the observed object trajectories\cite{salzmann2020trajectron++,nayakanti2022wayformer}, flow\cite{mahjourian2022occupancy}, or 4D occupancy\cite{zheng2024occworld,yang2025driving}. Detailed reviews are available in \cite{huang2022survey,kong20253d} and agentic mobility prediction is discussed separately in Sec. III.3.2.3. 
Here, we focus on completing the recovered 4D distribution at the scene level, preserving identity and geometry while jointly completing scene appearance and evolution. Object trajectories, flows or occupancies form one part of the underlying dynamics.  

\textit{(a) Video re-trajectory} is a related task rather than completion by itself. It re-renders input video along a new camera path without necessarily constructing a persistent dynamic field\cite{yu2025trajectorycrafter}. Nevertheless, its camera conditioning mechanisms\cite{bahmani2025ac3d,park2026redirector,wang2026bullettime} and synthetic multiview video training data\cite{bai2025recammaster,bahmani2025ac3d} provide useful priors for separating camera motion from native object dynamics.

\textit{(b) Video Warp-and-inpaint.} A typical 4D completion pipeline first renders partial observations along a novel camera trajectory, uses video diffusion to synthesize the missing evidence, and then reconstructs or fuses the completed observations back into the scene\cite{chen2026reconstruct}. Repeating this process progressively extends persistent 4D carriers through warping, inpainting, and final denoising\cite{liu2025free4d,lu2026see4d,gupta2024paintscene4d}. Driving systems further introduce dynamic point or LiDAR as conditions to preserve motion and consistency in completion\cite{wang2024stag,yan2025streetcrafter}. Across these variants, the central challenge is preventing iterative completion from accumulating geometric drift, distortion, and identity inconsistency\cite{huang2025vivid4d}.

\subsection{II.2 Urban Representation} 
Urban representation turns reconstructed urban states into an operable twin through two complementary tasks. Physical representation organizes geometry, appearance, and dynamics into persistent and updatable carriers, while information embedding grounds perceptual features and external urban knowledge on these carriers.

\subsubsection{II.2.1 Physical Representation} 
\label{sec:physical_representation}
Physical representations use mathematical models or structures to organize reconstructed geometry, appearance, and dynamics for rendering, querying, and interaction. Here we introduce how these representations are constructed and updated.

\begin{figure*}[t]
  \centering
  \includegraphics[width=\textwidth]{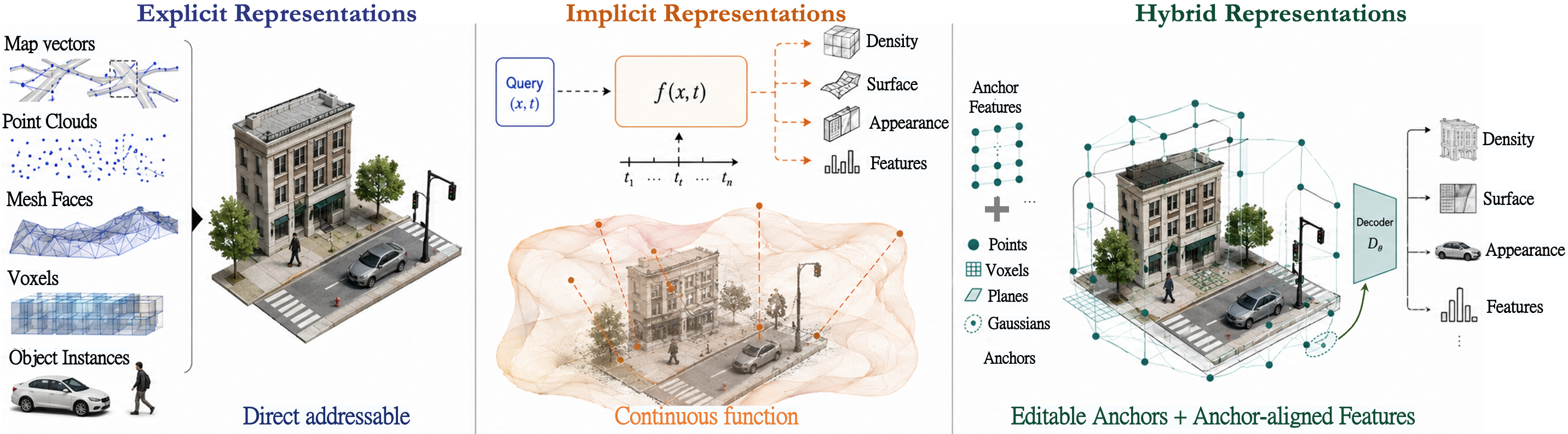}
  \caption{Three forms of physical representation. Explicit representations enumerate directly addressable primitives, implicit representations encode continuous scene functions, and hybrid representations attach learned features or attributes to explicit spatial anchors.}
  \label{fig:physical_representation}
\end{figure*}

\subsubsubsection{II.2.1.1 Representation Construction}
Based on how scene information is stored, physical representations of urban twin can be divided into the explicit, implicit, and hybrid forms illustrated in Fig.~\ref{fig:physical_representation}.

\subsubsubsubsection{Explicit Representations} Explicit representations store scene information in predefined geometric primitives, such as maps, points, and objects. 
At the early stage, vector structures such as OpenStreetMap\cite{haklay2008openstreetmap}, HD maps\cite{elghazaly2023high}, and scene graphs\cite{armeni20193d,fischer2024multi} mainly encode urban entities and their relations by vectors or records. Raster 2.5D products, such as digital elevation models\cite{shan2005urban} and digital surface models\cite{stal2013airborne}, support large area terrain and surface analysis. Standardized city modeling formats, such as CityGML\cite{groger2012ogc} and IFC\cite{venugopal2015ontology}, further define levels of detail for large scale cities\cite{chen2018ifc}. These abstractions support efficient query and record linkage, but often omit fine geometry, appearance, and temporal variation.

For higher fidelity, point clouds\cite{schonberger2016structure,yang2024ubiquitous} densely sample surfaces but lack explicit topology. Meshes\cite{chen2025meshanything,zhou2026mesh} support rendering and simulation by defining surface topologies but remain sensitive to holes, irregular topology, and reconstruction noise. Voxels and sparse grids\cite{kong20253d,contributors2022spconv,choy20194d} facilitate learning and volumetric computation but discretize continuous space. Object-level CAD and BIM elements provide structured entities, yet offer limited coverage of open world objects and visual variation\cite{groger2012ogc}. Overall, explicit representations provide interpretable geometry and can describe temporal evolution as sequences\cite{wang2024sequential}, but their predefined structures can become inflexible and costly for high fidelity appearance, frequent updates, and complex dynamics.

\subsubsubsubsection{Implicit Representations} Implicit representations encode scene information as learnable and structural functions such as MLPs or compact latent codes instead of enumerating geometric elements. Neural radiance fields\cite{mildenhall2021nerf,barron2021mip,tancik2022block,turki2022mega} represent a scene as a MLP that maps 3D positions and viewing directions to density and view dependent radiance, enabling photorealistic novel view synthesis through differentiable volume rendering. Surface neural fields instead map coordinates to signed distance or occupancy\cite{wang2021neus,peng2020convolutional,sitzmann2020implicit}. 
Extending the query domain from space to spacetime further enables implicit fields to model dynamic geometry and appearance\cite{pumarola2021d,yan2023nerf,zhong20243d}.
These lightweight formulations represent high fidelity geometry, appearance, and features continuously.

The cost is the expensive scene specific optimization, especially for the large urban scenes. For example, Block-NeRF\cite{tancik2022block} reports over 9 hours on 32 TPU v3 cores to reconstruct a single $100\,\mathrm{m}\times100\,\mathrm{m}$ street segment. 
Besides acceleration techniques\cite{garbin2021fastnerf}, some recent neural scene encoders use feed-forward scene latents to represent the scene from multiview inputs, replacing the per-scene optimization with learned priors\cite{jin2024lvsm,jiang2025rayzer,wei2026any}. These fixed-length latents serve as compact scene proxies that can be decoded efficiently into novel views\cite{jiang2025rayzer} or geometric structures\cite{wei2026any}. 

Implicit representations can also easily encode semantics\cite{zhi2021place,kerr2023lerf}, material attributes\cite{liu2023nero}, and additional sensing modalities such as infrared signals\cite{ye2024thermal}. However, such implicit forms reduce interpretability, making local editing less direct than with explicit representations.

\subsubsubsubsection{Hybrid Representations} Hybrid representations store scene information in explicit spatial anchors with learnable features or attributes. 

Compared with purely explicit forms, they allow a smaller or coarser set of anchors to embed and carry richer scene information. Typical anchors include points\cite{xu2022point}, sparse hierarchies\cite{zhong2023shine}, sparse grids\cite{muller2022instant,brown2025alphaearth}, or planes\cite{chen2022tensorf}. 
Gaussian Splatting (GS) has recently emerged as an important hybrid representation. 3D GS\cite{kerbl20233d} represents a scene with explicit anisotropic Gaussian kernels carrying optimizable position, covariance, opacity, and appearance attributes for fast visual rendering. For improved geometry, methods align Gaussians with implicit surfaces\cite{pan2025pings} or replace the kernels with oriented planar disks\cite{huang20242d}. For dynamic scenes, methods adopt native 4D Gaussians for spatio-temporal modeling\cite{yang2024real} or attach deformations\cite{wu20244d,xu2024grid4d} or trajectories\cite{wang2025shape} to the 3D primitives.

The explicit anchors enable efficient local access and editing comparing with purely implicit forms, without losing the learned part to recover continuous geometry, appearance, and semantics\cite{guo2026semantic}. 
However, when apply such good properties to urbans, the main limitation is also the scale. Naive 3DGS may contain millions or billions of Gaussian primitives to represent cities\cite{fang2024mini}, yielding slow training and rendering. We thus need techniques like distributed optimization\cite{zhao2025scaling,chen2024dogs}, sparsification\cite{fang2024mini,mallick2024taming}, hierarchical organization\cite{ren2024octree,lu2024scaffold}, and software or hardware acceleration\cite{ye2025gsplat,lee2024gscore,feng2025flashgs,ren2026fastgs}.

\subsubsubsection{II.2.1.2 Representation Updating}
Unlike a 4D representation that models motion within an observed interval, long term updating reconciles separately acquired sessions as the physical city changes. This process involves two related problems.

\subsubsubsubsection{Change Detection} Change detection compares cross-temporal observations to locate and classify additions, removals, and other persistent changes before updating the carrier. 
Typical methods construct handcrafted\cite{tran2018integrated} or learned features\cite{de2023siamese} for registered multi-temporal images\cite{zheng2024segment}, point clouds\cite{tran2018integrated,wang2023end}, or other observations, and predict change masks or semantic change classes through nearest neighbor comparison\cite{wang2023end}, siamese or latent space comparison\cite{de2023siamese,zheng2024segment}, or learned cross-temporal prediction\cite{tran2018integrated,zhan2024pgn3dcd,zhang2026me}. For broader coverage of 3D urban change detection, see \cite{stilla2023change}.

\subsubsubsubsection{Map Updating} Map updating incorporates confirmed changes into the scene carrier. For explicit representations, lifelong LiDAR mapping\cite{kim2022lt,yang2024lifelong} aligns point cloud maps acquired across sessions, identifies positive and negative changes, and selectively edits the corresponding geometry.
For implicit and hybrid representations, scene editing has advanced rapidly in small scale settings. Methods first localize target regions or entities through segmentation\cite{ye2024gaussian} and cross-time association\cite{ren2026cubifygs}, then support instruction modification and controlled deformation\cite{haque2023instruct,chen2024gaussianeditor,wu2024gaussctrl,tong2025cage}. These capabilities show potential for city scale updating; a broader taxonomy is available in \cite{zhu2026survey}.

\subsubsection{II.2.2 Information Embedding}
\begin{figure*}[t]
  \centering
  \includegraphics[width=\textwidth]{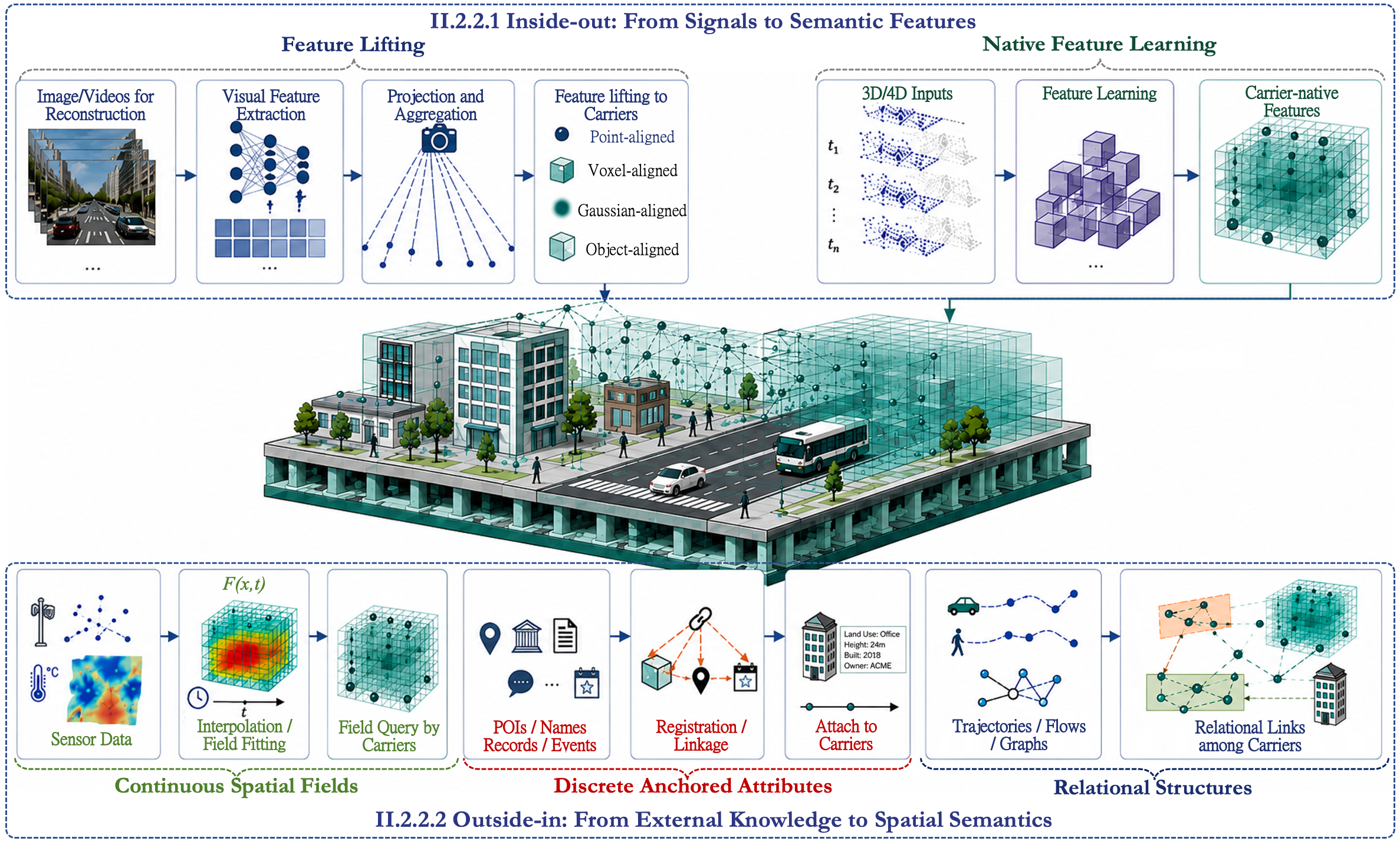}
  \caption{Strategies for embedding information in an urban twin. Inside-out methods lift or learn semantic features from reconstruction signals, whereas outside-in methods ground external fields, attributes, and relations on twin carriers.}
  \label{fig:information_embedding}
\end{figure*}

Information embedding operates on top of the reconstructed physical twin, augmenting its carriers with semantic features, external records, and relations. \textit{Inside-out} features are extracted from the data, such as images, videos, and point clouds, used to reconstruct the physical scene itself, whereas \textit{outside-in} information is introduced from auxiliary observations and external knowledge through spatial registration, entity linkage, or field construction. Fig.~\ref{fig:information_embedding} illustrates how both directions enrich the same carriers.

\subsubsubsection{II.2.2.1 Inside-out: From Signals to Semantic Features}
Inside-out embedding extracts semantic features from the reconstruction data to insert/link to corresponding carriers. Because task-specific features are better covered by reviews of their respective objectives\cite{guo2019deep}, we focus on reusable feature interfaces for downstream tasks. The discussion progresses from image and video features to representations lifted onto or learned natively from persistent 3D/4D carriers.

\subsubsubsubsection{Image Features} Large scale image pretraining produces complementary general purpose representations. Vision-language alignment equips CLIP\cite{radford2021learning} and SigLIP\cite{zhai2023sigmoid} features with open vocabulary semantics for categories and attributes. Mask supervision enables SAM\cite{kirillov2023segment} to delineate promptable regions and object boundaries. Self-distillation allows DINOv2\cite{oquab2023dinov2} to capture dense correspondences across visual variations. Masked reconstruction and latent prediction further encode contextual regularities in scene layout and object organization, as demonstrated by MAE\cite{he2022masked}, I-JEPA\cite{assran2023self}, and LingBot Vision\cite{fu2026vision}. VAE-based learning, in turn, constructs generative latent spaces that compactly encode image structure and semantic content\cite{chen2025comprehensive}. At Earth scales, AlphaEarth\cite{brown2025alphaearth} and SkySense++\cite{wu2025semantic} extend representation learning to multimodal satellite observations. Because these representations are not tied to fixed label sets, they transfer across diverse scenes and visual conditions.

\subsubsubsubsection{Video Features} Video features extend appearance and semantics with temporal structure. Video-language pretraining introduces transferable open-vocabulary semantics to video features\cite{wang2024internvideo2,zhao2024videoprism,bolya2026perception}. Masked reconstruction learns spatio-temporal context by recovering pixels, as in VideoMAE and VideoMAE V2\cite{tong2022videomae,wang2023videomae}, whereas V-JEPA and V-JEPA 2\cite{bardes2024revisiting,assran2025v} predict masked video latent features to capture underlying temporal dependencies and physical properties. Generative pretraining uses video tokenization and latent compression to learn compact spatio-temporal representations that preserve the structure required for reconstruction and synthesis\cite{yu2024language,wang2024omnitokenizer,brooks2024video,yang2025cogvideox,wan2025wan}.

\subsubsubsubsection{Features Lifting} 
Despite the strong representational capacity, the above image and video features remain view-centric, lying in image planes or structural latent spaces rather than physical coordinates. 
Information embedding must therefore align and lift them onto 3D/4D carriers, progressively moving from points to instances.

\textit{(a) Point-level features} associate each scene carriers with a visual or language embedding. ConceptFusion\cite{jatavallabhula2023conceptfusion} and OpenScene\cite{peng2023openscene} directly project and aggregate pretrained image features at 3D points. LERF\cite{kerr2023lerf}, LangSplat\cite{qin2024langsplat}, and GAGS\cite{peng2026gags} instead distill 2D language features onto radiance fields or Gaussian primitives for open-vocabulary querying. 4D LangSplat\cite{li20254d} extends this approach to 4D by supervising dynamic Gaussian fields with video descriptions.

\textit{(b) Instance-level features} lift multiview 2D masks into 3D to preserve object identity and boundaries. 
Early methods like SAM3D\cite{yang2023sam3d}, SAI3D\cite{yin2024sai3d}, and MaskClustering\cite{yan2024maskclustering} directly project 2D SAM masks onto point clouds for 3D instance masks by mask merging or clustering. 
Later methods replace explicit mask merging with learnable instance embeddings on differentiable scene carriers. For instance, Gaussian Grouping\cite{ye2024gaussian}, SAGA\cite{cen2025segment}, and Unified-Lift\cite{zhu2025rethinking} use rendered 2D SAM masks to supervise identity or affinity features on Gaussian primitives through 3D consistency regularization, contrastive objectives, or object level codebooks. LEGO\cite{peng2026lego} further lifts SAM masks into a 3D consistent hierarchy for levelwise Gaussian identity feature distillation.
These instances can be organized into scene graphs by assigning LLM-inferred relations\cite{gu2024conceptgraphs,zhang2025open,peng2026lego}, or tracked across time to form 4D graphs\cite{yang20234d}.

\subsubsubsubsection{Native 3D/4D Features} 
Besides lifting, some methods directly learn carrier-aligned features explicitly from 3D/4D carriers such as point clouds. 
This paradigm first matured on object data. Masked modeling learns transferable shape features\cite{yu2022point,pang2023masked}, VAE learning builds structural latents ready for generation\cite{xiang2025structured}, while multimodal alignment connects them with image and language semantics\cite{xue2023ulip,zhou2024uni3d}. 

At the scene level, native 3D representations, most commonly learned from point clouds, must additionally encode spatial context, heterogeneous scales, and relations. With scalable architectures such as sparse convolution\cite{contributors2022spconv} and point transformers\cite{wu2024point}, native 3D feature learning adopts contrastive learning and self-distillation to capture spatial context\cite{zhang2021self,hou2021exploring,wu2023masked,wu2025sonata}; masked point or voxel modeling to describe geometric structure\cite{hess2023masked,abdelsamad2025multi}; instance-aware objectives to preserve object separation and local organization\cite{yang2026towards}; VAE-based learning to construct generative latents\cite{guo2026scendi,hua2025sat2city,qian2026sat3dgen}; and cross-modal distillation to introduce open-vocabulary semantics\cite{huang2023ponder,chen2023clip2scene,peng2023openscene,zhang2026concerto}. Scaling these objectives across heterogeneous indoor, outdoor, and object data increasingly moves native 3D features toward a shared cross-domain space\cite{zhang2026utonia}.

For dynamic 4D observations such as point cloud sequences, existing methods learn features that remain invariant across temporally correlated frames\cite{huang2021spatio}, preserve object identity across tracked point clusters\cite{wu2023spatiotemporal}, or jointly encode shape and motion by reconstructing masked point tubes or features\cite{shen2023masked,zhang2023complete,zuo2025uni4d}. These representations remain concentrated on particular RGBD, LiDAR, or action datasets, leaving general scene level 4D features an open problem.

\subsubsubsection{II.2.2.2 Outside-in: From External Knowledge to Spatial Semantics}

Outside-in embedding grounds observations and knowledge that are not contained in the primary reconstruction signals in the physical twin. According to their spatial support, these data take three main forms, specifically continuous spatial fields, discrete anchored attributes, and relational structures such as trajectories, flows, and graphs.

\subsubsubsubsection{Continuous Spatial Fields} Continuous spatial fields provide values throughout the spatial or spatio-temporal domain required by the twin. Dense geographic observations that already cover the twin, such as registered thermal imagery, can be encoded directly as feature fields\cite{lu2024thermalgaussian}. Other georeferenced data provide only sparse samples. These may be physical measurements, such as temperature and air quality from sparse sensors\cite{li2024hyper,ahmad2025spatiotemporal}, or society observations derived from administrative and social sensing data, such as crime risk and urban sentiment\cite{hu2018spatio,gao2022measuring}. Both can be expanded into continuous fields with techniques like interpolation, spatial aggregation, or learned sparse-to-dense models\cite{comber2019spatial,li2023big}. The resulting field $\mathbf{F}(\mathbf{x},t)$ can be queried directly by location $\mathbf{x}$ and time $t$, or sampled at relevant twin carriers and stored as their attributes, in text or features.

\subsubsubsubsection{Discrete Anchored Attributes} Discrete anchored attributes attach information to identifiable objects, regions, events, or time intervals rather than defining values throughout continuous space. POIs and place names\cite{sun2023conflating}, descriptions\cite{chen2026spatialllm}, land use and functional zone labels\cite{yao2017sensing,chen2024urban}, administrative records\cite{haklay2008openstreetmap}, and event text\cite{anonymous2026lifeplanner} can all be retained as external textual records or encoded into features\cite{sun2025flexireg} and linked to their corresponding regions.

\subsubsubsubsection{Relational Structures} Relational structures preserve dependencies that cannot be reduced to independent locations or entities. Trajectories retain the ordered locations and states of individual entities\cite{parent2013semantic,feng2018deepmove}; flows aggregate movement between spatial units\cite{liu2021urban}; and spatial or knowledge graphs encode connectivity and high order relations among urban elements\cite{jiao2025survey,liu2023urbankg,ning2023uukg}. These structures may be linked to physical carriers through identifiers or encoded as trajectory-, node-, and edge-level features for message passing\cite{jin2025urban,sun2026urbanverse}. Broader taxonomies are available in related reviews\cite{parent2013semantic,jiao2025survey}.

\subsubsubsection{II.2.2.3 Continual Information Updating}
As new data arrive, information associated with scene carriers must be updated in coordination with the physical representation. This is necessary both when unchanged carriers acquire revised attributes and when physical carrier changes, requiring their information to be changed or reassociated.

\noindent\textbf{Inside-Out Updating.} Features and semantics extracted from new sensor observations are registered to their physical carriers. For a retained carrier, repeated labels or embeddings can be fused to refine its attributes through aggregation or merging\cite{mccormac2017semanticfusion,martins2025open}. When carriers physically change, however, changes to objects must be detected, moved or reassociated, and their associated features updated or replaced\cite{liu2025dynamem,jiang2025dualmap}.

\noindent\textbf{Outside-In Updating.} New external information is synchronized with the corresponding carriers.
Similar to the above embedding techniques, streaming sensor samples can continuously revise spatial fields through incremental interpolation or optimization\cite{liang2018real}.
Discrete records and events can be inserted, replaced, or invalidated through identifiers\cite{anonymous2026lifeplanner}. 
Relational structures require corresponding structural updates. Trajectories and flows are extended or revised as new states arrive, while graph edges are rewired when their nodes change\cite{hofmeister2024semantic}. Agentic construction can automate these operations at scale\cite{ning2024urbankgent,chen2026canglingknowflow}. Broader papers are reviewed in \cite{igelbrink2024online}.

Together, reconstruction and representation turn urban observations and external information into persistent, operable twin states. These states provide the foundation for the following urban agentic systems utilization.

\section{III. Urban Agentic Systems}

\begin{figure*}[t]
  \centering
  \includegraphics[width=\textwidth]{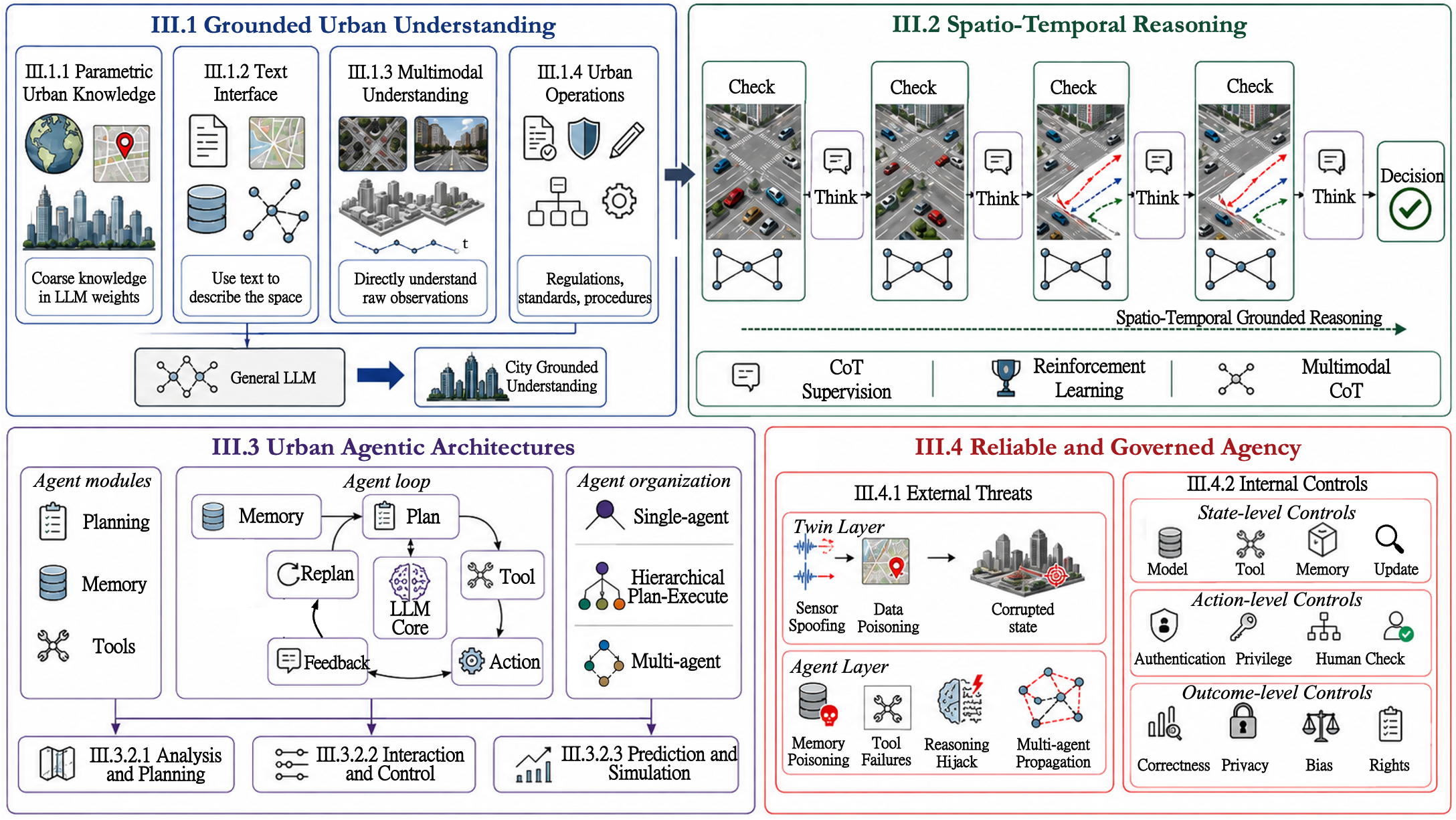}
  \caption{The four-layer organization of urban agent. General language models are grounded in city evidence, strengthened with spatio-temporal reasoning, embedded in agentic architectures, and constrained for reliable operations.}
  \label{fig:urban_agency_vision}
\end{figure*}

\noindent\textbf{A Vision for Urban Agent.} 
If the urban twin defines what city state is available, the urban agent determines how that state is accessed, interpreted, reasoned over, and acted upon. The key challenge is therefore not merely to build stronger LLMs or agents\cite{luo2025large,sang2025beyond}, but to ground their decision loop in the state constructed in Section~II.
We organize this discussion into four layers of urban agent, as illustrated in Fig.~\ref{fig:urban_agency_vision}.

Typically beginning with general LLMs, first, \textit{grounded urban understanding} determines how city-specific evidence become available to them by interfaces. Second, \textit{spatio-temporal reasoning} emphasizes whether entities, dynamics, and relations remain correct during the reasoning. Third, \textit{urban agentic architectures} place this reasoning core within a stateful loop of planning, memory, tools, execution feedback, and replanning, supporting urban analysis and planning, interaction and control, as well as prediction and simulation. Finally, \textit{reliable and governed agency} protects this loop from external manipulation and constrains its internal operations. The first two layers establish a city-grounded decision core; the latter two turn it into an accountable operational system. 

\subsection{III.1 Grounded Urban Understanding}
The key to grounded urban understanding is the interface through which agents read the urban twin\cite{zhang2024urban,zhang2025towards,hou2025urban}. As summarized above, twin states, including inside-out and outside-in information, are exposed in two forms, specifically textual records and features embeddings. 
In this section, we first discuss the drawbacks of LLM prior urban knowledge, and then review how textual and feature interfaces enable LLMs to understand urban space. In particular, we separately emphasize urban operational knowledge, which is essential for understanding how cities function.

\subsubsection{III.1.1 Parametric Urban Knowledge}
Parametric urban knowledge comprises geographic concepts and city information stored in LLM model weights\cite{manvi2024geollm,dorobantu2026geospatial}, without querying the twin.

\noindent\textbf{Inherent Urban Knowledge.} Pretraining on massive text corpora already gives LLMs a coarse ``virtual Earth''. GeoLLM\cite{manvi2024geollm} extracts these knowledge with coordinates and prompts constructed from map data, while CityLens\cite{liu2025citylens} finds that visual large language models can infer socioeconomic conditions from urban imagery. However, this awareness is biased across regions and scales. Geographic bias studies\cite{faisal2023geographic,manvi2024large} reveal systematic regional disparities, with higher factual errors in under-represented regions such as sub-Saharan Africa than in North America\cite{moayeri2024worldbench}. GPSBench\cite{truong2026gpsbench} further shows that performance declines from country-level to city-level knowledge.
Inherent parametric awareness is therefore a useful prior, yet remains insufficient for detailed city-scale analysis and planning\cite{zhao2025urban}.

\noindent\textbf{Knowledge Enhancement.} Some methods strengthen this implicit spatial knowledge through continued pre-training on geoscience text, as in K2\cite{deng2024k2} and GeoGalactica\cite{lin2023geogalactica}, or supervised fine-tuning on city-\cite{feng2025citygpt} and earth scale\cite{manvi2024geollm} tasks. Urban-R1\cite{wang2025urban} instead uses reinforcement learning to reduce cross-region disparities in urban profiling. These methods improve urban awareness, but their knowledge remains bounded by the training geography and is difficult to update\cite{dorobantu2026geospatial}.

\subsubsection{III.1.2 Text as an Intermediate Spatial Interface}

Textual interfaces expose twin states as language, records, graphs, or database query results\cite{chen2026spatialllm}. They can encode inside-out information by caption or description, as well as textual outside-in information, such as POIs, social records, and knowledge graphs.

\subsubsubsection{Indoor Scenes as Text} A common indoor pipeline\cite{gu2024conceptgraphs} first detects or segments objects from images or reconstructed 3D scenes\cite{ravi2025sam,carion2026sam}. It then converts object attributes and relations into text or text-aligned embeddings\cite{radford2021learning}, and organizes them as contexts or scene graphs. These textual representations can directly be read and selected by LLMs for scene understanding\cite{gu2024conceptgraphs}, object grounding\cite{yang2024llm}, navigation\cite{wu2024voronav}, and planning\cite{naren2024orionnav}. This approach is lightweight because it reduces a complex scene to a small set of objects and relations. It is also interpretable and easy to update\cite{ma2024llms}.

\subsubsubsection{Urban Space as Text} Some methods transfer these pipelines and advantages to urban space. However, cities contain many more entities, attributes, and relations, recorded across heterogeneous data modalities. Urban methods therefore address two additional problems: how to structure heterogeneous city data and how to retrieve only the evidence needed for a task.

Early work relied mainly on maps, which already abstract the city into objects and spatial relations\cite{haklay2008openstreetmap}, and focused on effective information retrieval. These approaches translate language requests into structured map database queries\cite{wan2026osmt}, coupling well-designed filtering strategies for more precise information grounding\cite{yu2026spatial}. 
Then, knowledge graphs provide a richer textual structure. For instance, urban state graphs connect physical urban entities and records\cite{liu2023urbankg,ning2023uukg,ning2024urbankgent}. Other graphs encode geospatial workflows\cite{liang2025geographrag}, possible causal relations among urban factors\cite{feng2025causal} to support LLM analysis.

Later work broadens textual information attached to maps and graphs. Historical trajectories supplement POI retrieval for travel planning\cite{ni2025tp}. LifePlanner\cite{anonymous2026lifeplanner} anchors noisy social media observations to OSM entities, while SpatialLLM\cite{chen2026spatialllm} attaches captions from imagery and point clouds to a map-based spatial skeleton. These interfaces make heterogeneous urban evidence searchable and traceable. 

However, text interfaces work well for textual information linked to a twin\cite{anonymous2026lifeplanner}, but cannot fully convey its appearance, metric geometry, motion, or latent features and fields\cite{jain2025words}. Moreover, general LLMs are proven to be weak at recovering spatial distributions from textual relations\cite{gardelakos2025can}, with performance on complex spatial analysis far below human levels\cite{zhao2025urban}.

\subsubsection{III.1.3 Multimodal Feature Interface}

Multimodal feature interfaces, beyond textual abstraction, directly expose detailed carrier features to the latent space of LLMs for direct spatio-temporal understanding. Existing studies mainly focus on inside-out features built from images/videos and 3D/4D carriers.

\subsubsubsection{Visual LLM Foundations}
The foundations of multimodal LLMs (MLLM) were established largely through research on image\cite{liu2023visual} and video\cite{li2024llavaone,li2024llavanext} understanding, termed Visual LLMs (VLM). 
A typical VLM uses a modality encoder to compress an input image or video to features, a lightweight adapter to map the visual features into the language space, and an LLM backbone to interpret the aligned multimodal content. 
Research has therefore progressed from pretraining visual encoders\cite{radford2021learning,zhai2023sigmoid,huang2024zero,hao2025urbanvlp} and designing lightweight adapters\cite{alayrac2022flamingo,li2023blip,dai2023instructblip,zhu2024minigpt}, to jointly tuning the adapter and LLM backbone with visual instructions\cite{huang2023language,hong2025glm,liu2024improved}. 
Recent work increasingly trains the above full architecture end-to-end\cite{bai2025qwen3,team2025kimi}. Native multimodal models further represent and process different modalities within a unified sequence\cite{wang2024emu3}, which enable multimodal understanding and generation by a single model\cite{hurst2024gpt,wu2025vila,team2024chameleon}. 
Although these advances make visual observations accessible to LLMs, they remain centered on 2D semantic alignment and still struggle with viewpoint changes\cite{kong2025lrr,zhou2025vlm4d}, metric scale\cite{chen2024spatialvlm}, and 3D relations\cite{liu2023visualspatial,liu2026abstract}.

\subsubsubsection{Geometry-Enhanced VLMs} A first response still preserves the image/video interface while introducing geometric supervision or position encoding, i.e., feature lifting. Specifically, VLM3\cite{cai2026vlm3} uses dense geometric tasks such as depth estimation, image matching, and camera pose objectives for geo-awareness supervision, while LLaVA-NeXT-Interleave\cite{li2024llavanext} introduces multiple images and 3D tasks like grounding to facilitate spatial understanding. SpatialRGPT\cite{cheng2024spatialrgpt} further combines region masks with RGBD metrics for 3D awareness. More directly, SpatialVLA\cite{qu2025spatialvla}, Video-3D LLM\cite{zheng2025video}, and GPT4Scene\cite{qi2026gpt4scene} provide the LLM with pixel-aligned 3D coordinates or spatially consistent labels as additional channels for spatial perception. Lmm-track4d\cite{li2026lmm} extends this approach with ray-time geometry embeddings for 4D awareness. However, these methods still divide a complete scene into separate multiview observations, limiting holistic scene understanding.

\subsubsubsection{3D/4D MLLMs} To avoid fragmenting a complete scene across multiple views, another line of work represents the scene directly in 3D or 4D for LLM processing. These methods encode spatial representations in different ways. They may use multiview lifting features\cite{fu2024scene,hong20233d,zhu2024llava}, 3D object features and scene graphs\cite{zhou2024uni3d,wang2023chat,huang2023chat}, native 3D scene features\cite{wei2026any,panagopoulou2024x,chen2024ll3da,mao2026spatiallm}, native 4D features extracted from dynamic scenes\cite{deng2024vg4d,sun2024l4d,zhou2026llava}. These features are then aligned with LLM latent space through an MLP\cite{hong20233d}, Q-Former\cite{tang2024minigpt}, or cross-attention adapter\cite{huang2023embodied,zhou2025uni4d} projection. 
Then, the LLM is trained for 3D perception like grounding, question answering, and spatial instruction following\cite{huang2023embodied,qi2024shapellm,chen2024ll3da,panagopoulou2024x}; 3D spatial analysis and planning\cite{li2026scot}; as well as unified 3D understanding, generation, and editing\cite{he2024llms,ye2025shapellm,ye2026universe3d}. 
In 4D space, it further supports dynamic tasks like 4D recognition\cite{deng2024vg4d,zhou2026llava}, tracking\cite{sun2024l4d}, dynamic scene graphs\cite{wu2025learning}, and 4D generation\cite{zhou2025uni4d}. However, most studies remain focused on indoor or small-scale scenes, whose object density, sensor diversity, and temporal updates are much simpler than those of a city.

\subsubsubsection{Urban-Scale MLLMs} Building on these architectures, urban-scale MLLMs process features from heterogeneous spatial data\cite{li2025urban}. These data include visual and geometric observations from LiDAR\cite{yang2025lidar}, Bird's Eye View (BEV) images\cite{ding2024holistic}, street maps\cite{venkatanarayanan2026mapreason}, city maps\cite{zheng2025urban}, multiview driving scenes\cite{zhuo2026drivetok}, etc.. They also include dynamic urban records, such as human trajectories\cite{gong2024mobility,ju2025trajllm} and socioeconomic activity over time\cite{jin2024time,yuan2024unist}.

Recent systems couple multiple urban data sources to move from basic perception toward advanced urban tasks\cite{zheng2025multimodal,hou2025urban}. OmniGeo\cite{yuan2025omnigeo} combines satellite imagery, geospatial metadata, and text for diverse GeoAI tasks. UrbanLLaVA\cite{feng2025urbanllava} integrates imagery, maps, and trajectories to reason across local and global urban views for tasks such as navigation and localization. CityAnchor\cite{li2025cityanchor} combines maps, images, and point clouds for coarse-to-fine object grounding, while 3DCity-LLM\cite{chen20263dcity} models objects, relations, and global scenes for scene analysis and planning. City-VLM\cite{sun2025city} fuses images and point clouds from various platforms for answering localization, measurement, functionality, and logicality problems. LifePlanner\cite{anonymous2026lifeplanner} instead grounds up-to-date socialmedia evidence in maps to support multi-target geospatial planning.

Multimodal training can ground LLM understanding in the spatial distributions and latent features or fields within an urban twin, offering substantial potential. However, this direction remains at an early stage. Existing datasets often cover only a few districts or cities\cite{feng2025urbanllava}, contain a limited range of urban modalities, and focus mainly on predefined perception, navigation, or spatial computing tasks\cite{chen2026spatialllm}. Scaling urban MLLMs therefore requires broader geographic range, richer multimodal data, and benchmarks that cover more open-ended urban tasks.

\subsubsection{III.1.4 Understanding Urban Operations}

Urban operational knowledge is a critical outside-in component, typically expressed as records or graphs extracted from regulations or other documents\cite{yang2025policy} and understood by textual interfaces.
Unlike observations that describe what is happening, it provides domain and up-to-date knowledge of how urban functions to general-purpose LLMs.

Early work focused on extracting and structuring it from regulations or government documents. For instance, OntoZoning\cite{silvennoinen2023semantic} represents zoning types, permitted uses, and programmes in an ontology linked to geospatial data, enabling cross-domain queries through a knowledge graph. Another study\cite{zhang2025large} extracts targets and actions from the Guangzhou Master Plan into a knowledge graph for efficient retrieval.

Recent systems ground LLM analysis in such knowledge. 
For instance, 
PlanGPT\cite{zhu2025plangpt} collects files like study materials, professional Q\&A, textbooks, government documents, and official plans. Retrieval or tuning with them supports LLMs on tasks like professional text generation, style transfer and information extraction. 
Planning chatbot\cite{zhou2025large} retrieves municipal bylaws and zoning regulations for accurate LLM-based regulatory queries.
ULLM-KG\cite{cai2026unified} designs domain knowledge graph construction, usage, and updating strategies to enhance LLM reasoning for fault diagnosis and maintenance recommendations in urban rail transit signal systems. 
DriveReg\cite{cai2026driving} retrieves traffic rules and guidelines relevant to the on-hand situation, then evaluates actions by legal compliance and safety with LLMs. 
GSI Agent\cite{wang2026gsi} builds a database from municipal manuals, planning documents, and inspection forms, supporting LLM fine-tuning, retrieval, and agentic usage for Green Stormwater Infrastructure tasks.

Recent works begin to extend the knowledge to visual space,
VLM-BCC\cite{zheng2025vision} retrieves building codes to generate inspection reports given building images.
VLM-Tunnel\cite{yu2025lightweight} detects tunnel defects with Qwen2.5-VL and retrieves technical standards with LightRAG to recommend maintenance actions.
VLM-RSIP\cite{fu2026unleashing} retrieves both technical standards and visual examples for attribute reasoning over roadside infrastructure images.

Operational knowledge extends LLM urban understanding from mere observations to the governing rules, which can also directly support subsequent professional reasoning and decision-making of LLMs\cite{wang2026gsi,cai2026driving,fu2026unleashing}. We therefore do not revisit these works in the next section.

\subsection{III.2 Spatio-Temporal Reasoning}

Basic perception makes urban evidence available, but does not ensure that subsequent LLM processing preserves it without distortion or error\cite{xu2025spatialbench,yang2025thinking}. Because complex urban tasks often require long reasoning chains, the intermediate steps must remain faithful to the underlying spatio-temporal scene states\cite{liang2025foundation,zheng2025multimodal}.

\subsubsection{III.2.1 Generic Reasoning}

The development of LLM reasoning follows a clear progression. Chain-of-Thought\cite{wei2022chain} (CoT) first elicited intermediate reasoning steps through prompting. Self-Consistency\cite{wang2022self} samples multiple CoTs and selects their most consistent answer, while Tree-of-Thoughts\cite{yao2023tree} (ToT) explicitly searches the best reasoning chain over alternative paths. Self-Refine\cite{madaan2023self} iteratively revises an output through self-feedback. Reflexion\cite{shinn2023reflexion} further converts feedback from previous attempts into experience for subsequent trials. Recently, training-based methods internalize reasoning through Supervised Fine-Tuning(SFT)\cite{zelikman2022star}, stepwise supervision\cite{lightman2024let}, and Reinforcement Learning(RL) with verifiable outcome rewards\cite{guo2025deepseek}. This development shifts reasoning from a prompted behavior toward an intrinsic capability of the language models.

VLMs extend this progression to think with visual inputs like images or videos\cite{li2025perception}. However, VSI-Bench\cite{yang2025thinking} shows that naive linguistic CoT, Self-Consistency, and ToT offer limited spatial tasks gains. For spatial tasks, the key point is to anchor reasoning process to real scene information, such as objects, coordinates, and relations. This motivates the discussions below.

\subsubsection{III.2.2 Spatio-Temporal Grounded Reasoning}

Spatio-temporal grounded reasoning remains at an early stage. It must preserve real scene distribution and motion in intermediate reasoning steps\cite{zheng2025multimodal}. Research progresses from indoor settings to robotic applications and early urban studies.

\subsubsubsection{III.2.2.1 Reasoning over 3D States} 3D reasoning maintains objects and relations as a persistent scene state. 

\subsubsubsubsection{CoT Supervision} 
For 3D object reasoning, 3D-CoT\cite{chen2025integrating} annotates 144k reasoning traces to teach LLMs connect shape and part geometry with function and causal effects. 
SCoT\cite{li2026scot} scales supervision to 1.1M scene-level samples and requires intermediate steps to correctly use scene evidence. Its results also show that grounded CoT benefits analysis and planning, whereas unnecessary reasoning can introduce hallucinations into basic perception.

\subsubsubsubsection{RL Enhancement} Later methods use reward design in reinforcement learning to encourage geometric reasoning with fewer CoT annotations. Spatial-MLLM\cite{wu2026spatial} combines 120k spatial QA data with only 2459 CoTs for cold start supervision, relying on GRPO-based RL for ability enhancement.
3D-R1\cite{huang20253d} generates its 30k cold-start CoTs and the GRPO stage rewards output format, 3D bounding boxes, and answer semantics. 
SpaceR\cite{ouyang2025spacer}, without annotating offline CoTs, asks the model to construct an explicit layout of scene objects during reasoning and rewards its agreement with ground truth maps. 
Scene-R1\cite{yuan2025scene} goes further with RL-only task adaptation that selects video evidence and reasons in the reconstructed scene.

\subsubsubsubsection{Robotic Testbeds} Robotics has served as an early testbed for such spatial reasoning\cite{wei2026agentic}. Supervised approaches decompose tasks into subgoal plans\cite{mu2023embodiedgpt}, ground reasoning steps in plans, motions and visual features\cite{zawalski2024robotic}, or predict future images as intermediate visual goals\cite{zhao2025cot}. Also, RL-based methods reduce trace annotation by rewarding verifiable keypoint states\cite{kim2026robot} and trajectory matches\cite{song2026maniplvm}. At a larger scale, Cosmos-R1 uses 1.93M reasoning traces and 30k RL samples to support general physical and embodied reasoning\cite{azzolini2025cosmos}.

\subsubsubsubsection{Urban Testbeds} Urban studies remain narrower and focus mainly on evidence based inference from urban imagery. CityRiSE\cite{liu2025cityrise} uses verifiable rewards in RL to identify socioeconomic cues in satellite and street view images, improving generalization to unseen cities and indicators. Urban-R1\cite{wang2025urban} uses urban region profiling as a proxy task and GRPO to learn more consistent reasoning across large scale geographic regions.

\subsubsubsection{III.2.2.2 Reasoning over 4D States} 
4D reasoning must update 3D spatial state through time. 

\subsubsubsubsection{Video Reasoning Foundations} General video methods provide an early foundation. VideoRFT\cite{wang2026videorft} shows that high quality CoTs and semantic consistency rewards can jointly improve video reasoning. MLLM-4D\cite{yin2026mllm} uses 2M basic QA to establish distance, direction, and motion awareness of input videos; 7k CoTs initialize reasoning; and 30k dynamic samples support RL that rewards agreement between intermediate reasoning and actual spatio-temporal states. Because textual traces cannot fully express spatial distributions, 4DThinker\cite{chen20264dthinker} interleaves language with visual latents in its multimodal CoT, using about 38k supervised and 37k RL samples.

\subsubsubsubsection{Ego-Centric Testbeds} Ego-centered systems such as robots and autonomous vehicles provide major testbeds for 4D reasoning\cite{wei2026agentic}. 
ST-Think\cite{wu2026st} combines long CoT supervision, reverse reasoning, and GRPO to understand dynamic scenes, 
whereas EgoReasoner\cite{zhu2026egoreasoner} emphasizes the reasoning templates and rewards should be task-specific. 
In driving, DriveLM\cite{sima2024drivelm} and DriveVLM\cite{tian2024drivevlm} structure reasoning from scene analysis to prediction and planning, while VLR-Driver\cite{kong2025vlr} recursively analyzes the risks and intentions of surrounding vehicles by spatiotemporal CoT. 
Moreover, RL-based methods\cite{jiang2025alphadrive,li2026drive,wang2025alpamayo} reward reasoning with trajectories and actions from different persepectives for driving control. Please refer to recent surveys\cite{jiang2025survey,kong20253d} for details.

\subsubsubsubsection{Urban Testbeds} Intrinsic reasoning over a continuously changing 3D city remains largely open. For example, TraveLLaMA\cite{chu2026travellama} builds 265k multimodal travel QA data, including 5k expert CoTs for spatial, temporal, and practical planning SFT. STReasoner\cite{ni2026streasoner} empowers LLM to integrate time series, graph structure, and text for explicit urban reasoning, with RL rewards designed for spatial information consistency. 

\subsection{III.3 Urban Agentic Architectures}

An urban agent supplies the operational intelligence for the environment represented by the urban twin, turning understanding and reasoning into action. This section outlines its architecture and applications in urban tasks\cite{han2025large}.

\subsubsection{III.3.1 From Reasoning Model to Agentic System}\ 

\noindent\textbf{Agent Modules.} Despite different taxonomies, agent surveys broadly converge on three core modules, specifically planning (reasoning), memory, and tool-mediated action\cite{sang2025beyond,han2025large}. 

\textit{(a) Planning.} Planning applies the reasoning capabilities discussed above to analysis the income, decompose goals, select actions, and replan from action feedback\cite{li2025review}. 

\textit{(b) Memory.} Agent memory is commonly divided by temporal scope: short-term (working) memory maintains the current task state and supporting evidence, whereas long-term memory retains factual knowledge and reusable experience across interactions\cite{qu2025comprehensive}.
Designing memory modules should consider what is retained to benefit reasoning, how it is encoded, and how it is retrieved and updated during execuation\cite{hu2025memory}.
More taxonomies further distinguish memory by function, form, and dynamics\cite{hu2025memory}.

\textit{(c) Tool.} Tools fullfill actions. They acquire evidence, perform specialist analysis, interact with environments, or validate results\cite{li2025review,luo2025large}. Tool use has progressed from fixed calls in handcrafted pipelines to autonomous selection or coding during reasoning and model native policies learned through post-training\cite{sang2025beyond}.

\noindent\textbf{Agent Loop.} Together, they form an ``Observe \(\rightarrow\) Retrieve Memory \(\rightarrow\) Plan \(\rightarrow\) Tool Execution \(\rightarrow\) Receive Feedback \(\rightarrow\) Update Memory'' \textit{loop} to resolve a complex problem \cite{li2025review,luo2025large}.

\noindent\textbf{Agent Organization.} At the implementation level, the same loop may be executed by a single agent, divided between planner and executor roles, or distributed across multiple specialists\cite{luo2025large,wei2026agentic}. Beyond this organizational choice, recent systems increasingly rely on a runtime harness that manages tools, persistent state, and execution lifecycles, providing the engineering infrastructure around the execution loop\cite{meng2026agent}.

\subsubsection{III.3.2 Agent Systems for Urban Space}

\begin{figure*}[t]
  \centering
  \includegraphics[width=\textwidth]{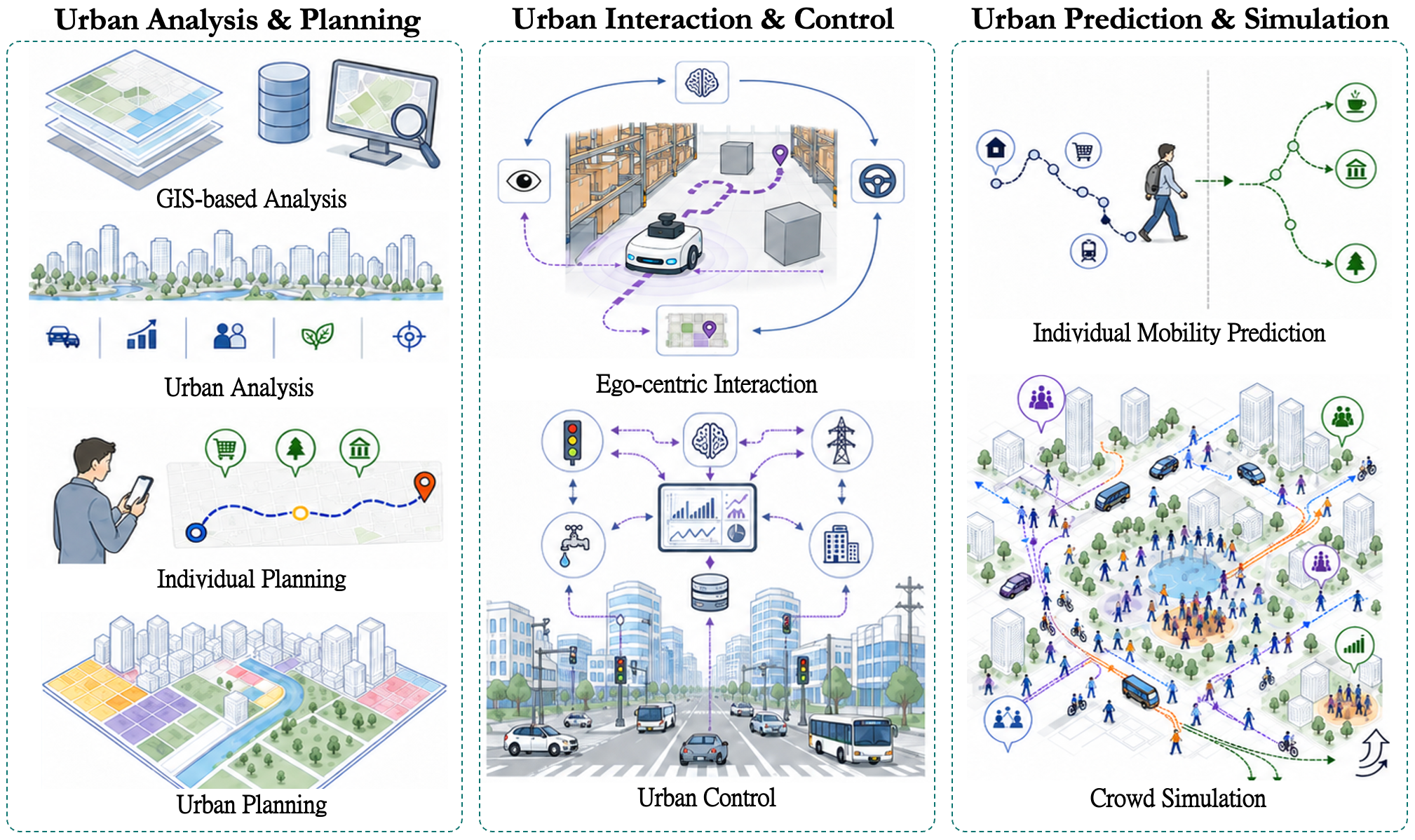}
  \caption{Urban agent systems organized by task. Analysis and planning transform urban evidence into analytical results and plans; interaction and control apply decisions to environments; mobility prediction and urban simulation model individual futures and collective dynamics.}
  \label{fig:urban_agent_tasks}
\end{figure*}

Urban agent systems apply the above agentic loops, with similar techniques, in different domains\cite{  zhang2025metacity}. We thus group them by tasks, specifically analysis and planning, interaction and control, as well as prediction and simulation, as summarized in Fig.~\ref{fig:urban_agent_tasks}.

\subsubsubsection{III.3.2.1 Urban Analysis and Planning} 
Agents here translate urban evidence into analytical results and planning proposals.

\subsubsubsubsection{GIS-Based Analysis} Early work used geospatial databases such as OSM\cite{haklay2008openstreetmap} and individual GIS tools as well-defined testbeds. 
OSMT\cite{wan2026osmt} translates natural language requests into executable OSM queries, while GeoTool-GPT\cite{wei2025geotool} teaches LLMs to use GIS analysis tools. Later systems integrate them into a complete GIS workflow. Autonomous GIS\cite{li2023autonomous} defines the GIS processing paradigm with five subgoals; GeoGPT\cite{zhang2024geogpt} coordinates data collection, processing, and analysis stream; MapGPT\cite{zhang2024mapgpt} follows a ReAct loop to operate cartographic tools to produce maps; and GeoFlow\cite{bhattaram2025geoflow} represents tasks as reusable workflow graphs.
Recent work further strengthens the workflow through autonomous processing code generation\cite{wu2025geocolab}, multi-agent cooperation\cite{mansourian2026bridging,lin2026geoagent}, as well as full-stack designs\cite{han2026gisclaw}. CangLing-KnowFlow further integrates an expert-validated procedural knowledge base with dynamic workflow adjustment and evolutionary memory, enabling end-to-end remote-sensing workflows to recover from execution failures and improve through accumulated experience \cite{chen2026canglingknowflow}. Complementing these runtime-oriented systems, SimCRAFT distills remote-sensing workflow-planning capabilities into compact 7B-scale models through constraint-validated synthetic tool-use trajectories and contextual retrieval-augmented fine-tuning, reducing dependence on large closed-source LLMs \cite{wang2026simcraftdistillingremotesensing}.

\subsubsubsubsection{Urban Analysis} Beyond fixed GIS databases, LLM-Find\cite{ning2025autonomous} enables agents to autonomously select and retrieve heterogeneous geospatial resources. Recent participatory urban sensing treats users as collaborative data collectors, adapting collection assignments to dynamic urban conditions and worker preferences\cite{guo2026agentsense} under fairness constraints\cite{guo2026language}.
These diverse data sources support a wide range of city-scale analyses. Representative examples include the following.

\textit{(a) Traffic analysis.} TrafficGPT\cite{zhang2024trafficgpt} decomposes complex questions and invokes traffic databases and foundation models for data processing, analysis, and interaction for advanced traffic management. TransitGPT\cite{devunuri2025transitgpt} generates and executes code for traffic data retrieval, calculations, and visualization.

\textit{(b) Urban change analysis.} ChangeGPT\cite{xiao2025llm} calls visual foundation models for change perception, whereas MMUEChange\cite{xiao2026mmuechange} uses a tool library and modality controller to align remote sensing, geographic, and socioeconomic evidence for analyzing changes in park construction, water quality, and waste site distributions.

\textit{(c) Socioeconomic analysis.} Beyond spatial distributions, agents can also derive socioeconomic indicators. SAGAI\cite{perez2025streetscape} combines OSM and street view imagery to derive street indicators such as storefront presence and sidewalk width, while UrbanAgent\cite{hao2026multi} resolves cross-modal conflicts by coordinating specialist agents that actively retrieve evidence for region level carbon emission, GDP, and population inference.

\textit{(d) Environmental analysis.} UrbanSoil\cite{qin2026toward}, for instance, links parcels, land uses, pollutants, measurements, pathways, and regulatory thresholds to corresponding geospatial entries to guide professional environmental dialogues. DORA\cite{wang2026can} requires agents to combine 108 geospatial tools over heterogeneous geospatial data for disaster analysis, rescue and evacuation planning, and reporting over time.

\subsubsubsubsection{Individual Planning} The above urban analyses provide direct support for planning at both the user and city scales. At the individual level, ITINERA\cite{tang2024itinera} decomposes user requests, retrieves POIs, and applies cluster spatial optimization to generate itineraries. TraveLLaMA\cite{chu2026travellama} uses reasoning, tools, and reflection for interactive travel assistance. DiMA\cite{ning2025dima} brings this pattern into a specific ride-hailing service, where tools support progressive order planning under cost constraints. With planning tasks of varying difficulty, LifePlanner\cite{anonymous2026lifeplanner} shows that noisy evidence retrieval, tool use failure, and balancing multiple constraints remain key bottlenecks for agentic planning.

\subsubsubsubsection{Urban Planning} For cities, existing work mainly focuses on land use planning. For instance, OntoZoning\cite{silvennoinen2023semantic} links permitted land uses and programmes under each zoning type for urban planning in Singapore. LLLM-PUP\cite{zhou2024large} lets a planner propose land uses, collect feedback from resident agents, and revise the proposal. LiPUP-MA\cite{ni2024lipup} extends this loop by letting residents experience a simulated plan and storing their feedback for revision. Intelli-Planner\cite{yong2026intelli} further introduces RL, using five evaluation dimensions to optimize plan generation.

However, urban agents still face drawbacks like domain constraints and regulatory hallucination\cite{yang2025reasoning,yang2025policy}. Generalizing agentic analysis and planning across urban scales and domains thus still requires stronger grounding in urban theory and planning legitimacy\cite{tong2026bridging}. 

\subsubsubsection{III.3.2.2 Urban Interaction and Control} Agents here execute actions within an environment and refine subsequent decisions through feedback. 

\subsubsubsubsection{Ego-Centric Interaction}
One class of systems controls an embodied ego agent, such as a robot or a road vehicle. Although embodiments and action spaces differ, these systems share the same mechanism that the agent grounds local observations into a representation of itself and its surroundings, selects an action, receives environmental feedback, and updates its memory or plan before acting again. 

General embodied tasks establish this pattern. Inner Monologue\cite{huang2022inner}, for example, interleaves language planning with motion and visual feedback, while DEPS\cite{wang2023describe} uses environmental feedback to re-describe the state, explain failures, and reselect plans for long horizon tasks.

Urban exploration extends this closed loop to city spaces. 
VELMA\cite{schumann2024velma} continually incorporates street view descriptions and trajectory history before choosing the next navigation action. 
EmbodiedCity\cite{gao2024embodiedcity} enhances the simulated cities with pedestrian and vehicle flows, where agents perform scene understanding, question answering, dialogue, navigation, and task planning. 
CityNavAgent\cite{zhang2025citynavagent} combines hierarchical semantic planning with global topological memory for aerial navigation.
CityEQA\cite{zhao2025cityeqa} separates question decomposition, objectic map maintenance, and embodied exploration among a multi-agent system.

The widely studied autonomous driving systems apply the same loop under faster dynamics and stricter safety constraints. 
DiLu\cite{wen2024dilu} combines reasoning and reflection to make decision based on common-sense knowledge and accumulate experience from successive driving decisions. 
Agent-Driver\cite{mao2023language} further couples callable tools and cognitive memory with reasoning, task planning, motion planning, and self-reflection, thus enabling a more nuanced, human-like approach to autonomous driving. 
DriveAgent-R1\cite{zheng2026driveagent} strengthens feedback interpretation through depth estimation, detection, and POI tools over recent views. 
AgentsCoDriver\cite{hu2024agentscodriver} further supports collaborative driving through cognitive memory, reinforcement reflection, and communication, allowing vehicles to exchange information and accumulate experience through continued environment interaction.

\subsubsubsubsection{Urban Control} 
Unlike ego-centric systems that control a mobile embodied subject, these agents act directly on urban infrastructure and services.
A typical task is traffic signal control. LLMLight\cite{lai2025llmlight} prompts an LLM with realtime traffic conditions to reason and select signal phases. 
LLM-Assisted Light\cite{wang2024llm} further augments the LLM with perception and decision-making tools, connecting LLM reasoning capabilities with established control methods. 
Moving beyond a single light intersection, CoLLMLight\cite{yuan2026collmlight} constructs a structured spatiotemporal graph of realtime traffic dynamics and relationships among neighboring intersections, uses complexity-aware reasoning for network-wide light control. 
More broadly, TrafficClaw\cite{lai2026trafficclaw} models systems like traffic signals, freeways, public transit, and taxi in a unified traffic environment and combines spatiotemporal reasoning, persistent memory, and agentic RL for joint coordination.

Beyond traffic, smart building agents can perceive building energy use and manage connected appliances\cite{he2025context}; 
provide personalized building services such as lighting adjustment based on user profiles\cite{saleh2025usercentrix}; 
or combine LLM reasoning with lightweight physical models to estimate thermal comfort, energy use, and wall heat gain across a city block\cite{lai2026agentic}.
Grid-Agent\cite{zhang2025semantic} targets power grid control, where a planning agent uses power flow solvers to generate coordinated action sequences with a validation agent checks stability and safety through sandboxed execution. 
WaterAdmin\cite{wen2026wateradmin} focuses on community water systems, using an upper level LLM abstracts weather variations, human activities, and human instructions into operational targets, as well as a lower level optimizer controls pumps and valves.
However, most existing studies remain confined to a single urban system and rely on simulations, benchmark grids, or offline datasets. A substantial gap remains before agents can achieve safe, long-term, and autonomous coordination and control across domains in real urban systems.

\subsubsubsection{III.3.2.3 Urban Prediction and Simulation}
\label{sec:agentmotion}
These systems combine historical information with system design to predict or simulate future states of individuals or crowds.

\subsubsubsubsection{Individual Mobility Prediction and Generation} From an individual perspective, agentic systems mainly focus on next location prediction and trajectory generation by fullfilling the individual preferences\cite{liu2023can}. 
For the former, AgentMove\cite{feng2025agentmove} combines individual mobility pattern, urban distribution, and shared patterns to infer the next location from historical trajectories. 
ARMove\cite{wang2026armove} further uses multiple agents to iteratively optimize mobility, memory, and social features across regions and groups.
Trajectory generation instead constructs personalized, spatially grounded activity chains. 
LLM-UR\cite{wang2024large} and TrajLLM\cite{ju2025trajllm} characterize personal activity patterns before selecting activities and destinations to form trajectories. 
CAMS\cite{du2025cams} further combines individual mobility patterns with urban spatial information and collective mobility distributions.
Recent TrajAgent\cite{du2026trajagent} uses LLM agents to select and optimize specialized models within a unified framework, jointly supporting and optimizing tasks including trajectory prediction, recovery, classification, generation, and representation. 
Further advances are reviewed in~\cite{guo2025language}.

\subsubsubsubsection{Multi-Agent Crowd Simulation} From an urban perspective, multiple agents typically interact repeatedly with a shared urban environment to simulate future city states\cite{ye2026towards}. Generative Agents\cite{park2023generative} established the basic architecture, showing that persistent memory and autonomous interaction can produce emergent social behavior. Building on this architecture, MobileCity\cite{ye2026mobilecity} grounds the decisions of 4k agents in evolving principles such as needs, habits, and obligations, producing more human-like crowd mobility. AgentSociety\cite{piao2025agentsociety} instead emphasizes scale, simulating more than 10K agents and 5M interactions to examine the patterns, causes, and mechanisms of five social issues. CitySim\cite{bougie2025citysim} targets long-horizon simulation by endowing tens of thousands of agents with beliefs, long-term goals, and spatial memories for sustaining simulation.  MatrAIx\cite{li2026matraix} represents heterogeneous users through 8.3 billion records defined over 1290 categorical dimensions, while records are sampled as active agents only when paired with a model, interface, and task.

Other systems couple agents with specific urban environments. GATSim\cite{liu2026gatsim} embeds 70 agents with hierarchical memory in a transport simulator, where the interaction produces collective mobility and traffic patterns. RESPOND\cite{sultimov2026respond} couples population agents with realtime flood forecasts, enabling communication for joint evacuation and resource seeking as hazards evolving. 

As simulations expand in scale, duration, and environmental detail, their computational cost becomes a major bottleneck. OpenCity\cite{yan2024opencity} designs request scheduling and ``group-and-distill'' prompt optimization to achieve a 600-fold speedup in agent simulation time. CityBehavEx\cite{santos2026citybehavex} instead combines established mobility models with fine-tuned cross-encoders that estimate semantic alignment between agent profiles, schedules, and activity transitions, simulating 100K agents for 75 days in under one hour on a single consumer GPU. MatrAIx\cite{li2026matraix} addresses efficiency through selective participation and parallel execution, where each evaluation samples only a cohort required by a task among 8.3 billion candidates.

Scale, however, does not ensure empirical realism. 
MatrAIx shows that simulated outcomes can vary substantially with the underlying agent model, for instance ranging from 23.2\% to 93.9\%,  even with identical persona agents follow their well-designed attributes\cite{li2026matraix}. 
WPINR shows that large-scale urban simulators may capture some high-level semantic activity distributions while failing to reproduce core and fine-grid spatial and temporal constraints, such as dwell times and transition dynamics\cite{santos2026plausible}. 
Credible simulations of urban futures therefore depend on the joint fidelity of the simulated population, environment, and agent system. Only by combining these elements can simulated citywide dynamics more closely approximate reality.

\subsection{III.4 Reliable and Governed Agency}

Reliable urban agency depends on two boundaries: protection against external threats and internal controls that constrain system operation\cite{ma2026safety}.

\subsubsection{III.4.1 External Threats}

Urban agents combine retrieved documents, maps, sensor streams with user instructions, agent memory, tool responses, and inter-agent messages, allowing untrusted content to apply. 
At the twin layer, for instance, sensor spoofing and data poisoning can corrupt the observed state\cite{sato2025wip,lou2025asymmetry} to bias agent estimation and prediction\cite{wang2024data}. 
At the agent layer, attacks can poison memory, compromise tool use and protocols, hijack reasoning, or propagate through multi-agent communication. A comprehensive taxonomy of these attack surfaces and defenses is available in \cite{ma2026safety}.

\subsubsection{III.4.2 Internal Controls}

Beyond defending against attacks, internal reliability requires governing and constraining the agent throughout its operation. These constraints broadly apply at three levels.
First, state level controls govern what may be stored, retrieved, or modified in the memory, tools, and model\cite{hu2025memory}.
Second, action level controls constrain agent delegation and tool use through methods like authentication, least privilege, and human approval\cite{south2025authenticated}.
Third, outcome level controls screen plans and outputs for biased claims, privacy violations, and threats to fundamental human rights\cite{weidinger2022taxonomy}.
A broader review is available in \cite{zhou2026externalization,ai2024artificial}.

\section{IV. Future Vision}

Looking ahead, Section~IV.1 examines how the capabilities reviewed above can be integrated into a coherent system and identifies the technical advances required, while Section~IV.2 considers how this system prototype can be adapted across urban domains.

\subsection{IV.1 Future System Vision and Research Directions}

\subsubsection{IV.1.1 System Blueprint: Two Loops}

\begin{figure*}[t]
  \centering
  \includegraphics[width=\textwidth]{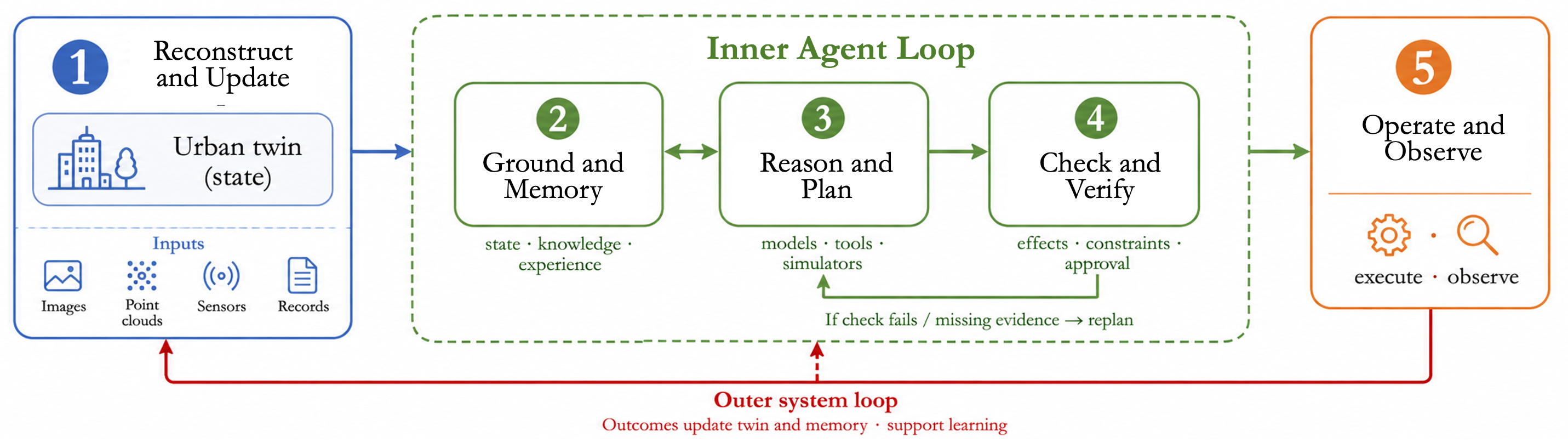}
  \caption{The proposed system blueprint. The inner loop grounds, reasons, and acts, then replans before action. The outer loop updates the twin, executes an approved decision, observes the outcome, and starts the next full cycle.}
  \label{fig:twin_agent_blueprint}
\end{figure*}

The methods reviewed above support the system blueprint in Fig.~\ref{fig:twin_agent_blueprint}.

The (1) \textit{Reconstruct and update} step uses new multimodal observations and external records to reconstruct the city and update its twin.
The updated twin supports the \textit{\textbf{inner agent loop}}: \textit{(2) Ground and memory} $\leftrightarrow$ \textit{(3) Reason and plan} $\leftrightarrow$ \textit{(4) Check and verify}. In Step~(2), the agent retrieves the current city state, relevant knowledge, and past experience for subsequent analysis. In Step~(3), it uses reasoning, tools, and simulators to plan, analysis or simulate an intervention. In Step~(4), it checks the simulated results, expected effects and constraints. Missing evidence or a failed check returns the agent to Step~(2)-(3) for memory updating and rethink.
A final validated agent decision moves from Step~(4) $\rightarrow$ \textit{(5) Operate and observe}. The observed decision outcome then returns from Step~(5) $\rightarrow$ Step~(1). This closes the \textit{\textbf{outer system loop}}. Each outcome updates the twin and agent memory. Outcomes collected over time can also improve the models, tools, and policies.
Steps~(2)-(4) therefore refine a decision before applying. Steps~(1)-(5) connect city sensing, decision, and learning. 
Both loops should be governed through auditable state management, bounded authority, and human oversight.

\subsubsection{IV.1.2 Research Directions}

The following directions move from improving individual components to designing, training, and governing the complete system.

\noindent\textbf{All-in-One Urban Twinning.} Urban twin construction still separates reconstruction, completion, representation, and information embedding, which limits joint optimization. Building on recent feed-forward methods\cite{liang2025wonderland,keetha2026mapanything,li2025iggt,zhu2025aether}, future models should quickly and accurately turn fragmented multimodal observations into a complete urban twin with both physical states and rich features.

\noindent\textbf{Urban Agent Loop.} Urban agent must move from narrow indoor tasks to complex city environments and applications. The agent should understand the large and diverse state stored in the twin, and its reasoning must remain grounded in urban information and knowledge. It should also coordinate multiple urban domains and tasks\cite{zhang2025metacity}, combine short and long memories at the city scale, and use tools for citywide analysis and simulation. The complete inner loop should improve these capabilities together.

\noindent\textbf{Twin-agent Interface.} Although urban twins and agents can be developed separately, they must operate and evolve as one system, making their interface critical. Current approaches connect independently designed twins and LLMs by textualizing twin states or projecting features through adapters. These post hoc bridges lose spatio-temporal detail and weaken grounding. Future systems should instead co-design twin representations and agent architectures, encoding aggregated urban states as agent-native tokens that agents can process directly. Training losses or online feedback could then back-propagate through the differentiable agent, interface and twin. Unified multimodal models\cite{wang2024emu3,wu2025vila} offer an early prototype by representing different modalities as token sequences for joint training, but they do not yet support persistent, heterogeneous, and dynamically updated urban twins.

\noindent\textbf{Urban Task Taxonomy.} Existing tasks, such as reconstruction, segmentation, question answering, and navigation, are mostly elementary or confined to specific scenarios, with limited connections and little systematic organization\cite{li2026scot}. Real urban problems instead compose multiple basic capabilities into higher level tasks, with requirements varying across scenarios and domains. Yet a coherent taxonomy of urban intelligence tasks is still missing, even within individual domains such as buildings or transportation. A broad and progressive task hierarchy, spanning basic capabilities to complex urban workflows, is essential for designing system architectures, annotating training data, and evaluating systems toward general spatio-temporal intelligence.

\noindent\textbf{Urban Data Annotation at Scale.} Beyond a task taxonomy, existing urban datasets remain limited in geographic coverage (often spanning only a few regions or cities) as well as in modality diversity, temporal depth, and annotation granularity\cite{anonymous2026lifeplanner,feng2025urbanllava,feng2025citygpt}. Future datasets should combine city-scale multimodal observations over time with operational knowledge, including rules and regulations. Annotations should align twin construction and updating, agent reasoning and action, and observed outcomes across diverse urban tasks. Such data would support component training and evaluation as well as end-to-end optimization of the full system.

\noindent\textbf{System Governance.} Governance must address both external threats and internal controls. External protection should defend sensors, data, memory, tools, and communication against manipulation. Internal controls should enforce versioned state, bounded authority, and rights-aware human oversight. Both forms of protection must remain active during training, deployment, and later updates.

\subsection{IV.2 From a Shared Prototype to Urban Domains}

The two-loop structure provides a shared twin-agent prototype rather than a domain-ready system. Each domain can specify the state maintained by the twin, the models and actions available to the agent, and the authority required for operation. The following examples show how these interfaces turn the shared prototype into domain-specific closed loops.

\noindent\textbf{Built Environment and Infrastructure.}
A built environment twin would align geometry with asset semantics and inspection histories. The agent could retrieve detected changes and applicable standards, analysis and reason, use domain models to assess the intermediate steps, and verify a maintenance plan. Approved work and subsequent inspections would close the outer loop by recording the actual outcome. Existing building agents provide early components for this workflow, which could extend to other infrastructure assets\cite{he2025context,lai2026agentic}.

\noindent\textbf{Traffic and Mobility.}
Here, the key of the twin combines road topology with current traffic and signal states. The agent could use forecasting and simulation tools to compare control strategies under network-level constraints. Approved decisions would then be deployed, while sensor observations return their effects to the twin. 
Early traffic loops\cite{chen2024toward}, simulators\cite{gu2026large}, agents\cite{yuan2026collmlight,lai2026trafficclaw} provide key components of such a close loop system.

\noindent\textbf{Environment and Urban Ecology.}
An environmental twin should link actors like land cover and vegetation with measured or inferred environmental fields. Grounded in this state, the agent could use ecological models to identify intervention areas and compare candidate strategies under feasibility constraints\cite{pan2024assessing,qin2026toward}. After implementation, remote sensing and field observations would update both the environmental state and the estimated effects of the intervention.

\noindent\textbf{Citizen Services and Public Safety.}
A citizen service twin should represent service demand and population state under privacy requirements. Its agent could perceive these information, adapt services or guide population flows after prediction or simulation checks\cite{li2026matraix,liu2026fusion}. During an emergency, the same loop could analysis evolving hazards to formulate responses, and update them as field observations envolve\cite{wang2026can,sultimov2026respond}.

\section{V. Conclusion}

This review has examined STI systems in urban space through a unified twin-agent perspective. For urban twinning, we reviewed the reconstruction and completion of urban states, followed by their physical representation and information embedding. For urban agent, we examined how grounded understanding and reasoning are incorporated into agentic architectures under reliability and governance mechanisms.

Building on these components, we presented a future STI system blueprint with an inner agent loop for grounding, reasoning, and verification, together with an outer system loop connecting urban sensing, action, and observation. We also summarized its research agenda and adaptation across urban domains. Together, these perspectives provide a unified account of how urban states can be constructed, interpreted, and connected to accountable decision making.

\section{VI. Acknowledgements}

This research is jointly supported by the General Program of NSFC (No.42571521) and PhD Program of NSFC (No.424B2012).

\section{VII. Declaration of interests}
The authors declare no competing interests.

\section{VIII. Declaration of generative AI and AI-assisted technologies in the writing process}
During the preparation of this work, the authors used ChatGPT in order to improve language and readability. After using this tool/service, the authors reviewed and edited the content as needed and take full responsibility for the content of the publication.

\nolinenumbers
\bibliographystyle{ACM-Reference-Format-surname-first}
\bibliography{reference}

\end{document}